\documentclass[11pt]{article}
\usepackage[margin=1in]{geometry}
\usepackage[utf8]{inputenc}
\usepackage[T1]{fontenc}
\usepackage{lmodern}
\usepackage{amsmath,amssymb}
\usepackage{booktabs}
\usepackage{graphicx}
\usepackage{xcolor}
\usepackage[expansion=false]{microtype}
\usepackage[round,authoryear]{natbib}
\usepackage{listings}
\usepackage{placeins}
\usepackage[hidelinks]{hyperref}
\newcommand{\Kone}{$K{=}1$}
\newcommand{\bptt}{BPTT}
\newcommand{\xdeg}{$x_{\mathrm{deg}}$}

\graphicspath{{figs/}}
\newcommand{\PTTrainingRuns}{1,092}
\newcommand{\PTTrainingHours}{75}
\newcommand{\PTPostPilotP}{0.004}

\begin{document}

\title{Fitting Every Hint Is Not Enough: Premature Floods in Recurrent Algorithmic Reasoners}
\author{Amin Najafgholizadeh\thanks{University of Zanjan. Emails: \texttt{amin.ngz@znu.ac.ir}, \texttt{afsharchi@znu.ac.ir}.} \and Mohsen Afsharchi\footnotemark[1]}
\date{}
\maketitle

\begin{abstract}
Recurrent algorithmic reasoners are trained with per-step hints on small graphs and run for many more steps on large ones. On breadth-first search in SALSA-CLRS, we find that a reasoner can reproduce every training hint and still fail on deep graphs: after some exact steps, it declares nodes reached before the search gets there. We call this a premature flood and trace two routes to it. In our GRU processor with a degree input, the flood is carried in the hidden state and set off when that input is in the tail of the training distribution. Setting the input higher at inference brings the flood sooner, and none of the six flooding models counts steps. Without the input, summing the neighbors' states instead of averaging them also breaks the processor. In the benchmark's own GIN and PGN, trained with its code, some seeds instead flood from a fixed step, on sparse and dense graphs alike. Clamping the degree input or adding 200 dense training graphs removes every failure of the GRU processor. Detaching the state between steps avoids the flood in that processor (0 of 23 seeds, against 6 under full backpropagation) but not with sums or a residual update, and two published remedies leave floods. None of the benchmark's test sets is both deep and dense, but its deepest one shows the second route: on its Delaunay graphs at 1,600 nodes, one GIN seed floods from step 18. We therefore recommend auditing the range of every input and aggregate, which catches the first route, and testing on deep graphs, which catches both.
\end{abstract}

\section{Introduction}
\label{sec:intro}

\begin{figure}[t]
\centering\includegraphics[width=0.62\linewidth]{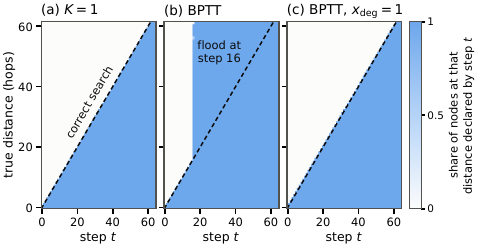}
\caption{One deep dense test graph (a blob of 64 cliques, 1,024 nodes). For every true BFS distance, each panel shows the share of the nodes at that distance declared reached by step $t$. The correct search fills the region below the dashed diagonal. (a) A reasoner trained with \Kone{} is exact. (b) The same seed trained with \bptt{} is exact until step 16, when it first declares nodes reached before the search arrives, and declares every node by step 17. (c) The weights of (b), with the degree input clamped to 1.0 at inference, are exact again. The model and the graph were chosen by a rule fixed in advance (Appendix~\ref{app:registrations}).}
\label{fig:teaser}
\end{figure}

A neural algorithmic reasoner is trained to execute an algorithm step by step \citep{velickovic2022clrs}. On graphs, a message-passing block carries a hidden state between steps and is supervised by the algorithm's intermediate state, the \emph{hint}. CLRS and SALSA-CLRS train on graphs of sixteen nodes or fewer \citep{velickovic2022clrs,minder2023salsaclrs}, then run the block for as many steps as a larger graph needs. Does fitting every training hint ensure that the learned update still works hundreds of hops deep?

For breadth-first search (BFS) the hint is the set of nodes reached so far, and the benchmark's training graphs are shallow: in our draw of them, 99\% are at most five hops deep. The hints therefore constrain the carried state for a few steps only, and only under inputs like the training ones, while a larger graph brings larger degrees and larger sums over neighbors. In CLRS a shift in degree alone did not harm BFS \citep{mahdavi2023ood}, but those test graphs were shallow. We find that a reasoner can fit every hint and still fail on graphs that are deep and dense: after some exact steps, its readout declares nodes reached before the search gets there (Figure~\ref{fig:teaser}). We call this a \emph{premature flood}. These timing errors can corrupt the decoded parent pointers even when the final reached set is complete.

Our main model, a GRU processor that reads each node's degree, is trained like every SALSA-CLRS baseline, with full backpropagation through its applications (\bptt{}). Under \bptt{}, 6 of 23 seeds flood on the deepest blobs. In this processor the flood is carried in the hidden state: resetting that state every four steps at inference, with the reach mask kept, repairs five of the six, and the sixth begins by step five (\S\ref{sec:mechanism}). It is set off by the degree input when the input lies in the tail of the training distribution. Raising the input at inference brings the flood sooner. A state could instead count steps and fire after a learned number of them; under an empirical onset criterion fixed in advance, none of the six flooding models qualifies. The readout acts on where the state has moved, and a tail input moves it (\S\ref{sec:drift}). Without the degree input, summing the neighbors' states instead of averaging them also breaks the processor (\S\ref{sec:sums}).

The second route belongs to the benchmark's own models. Its GIN and PGN, trained with its code, take no degree input, yet some of their seeds flood from a fixed step, on sparse mazes and dense blobs alike (\S\ref{sec:benchmark}). The route shows on the benchmark's deepest test set: on its Delaunay graphs at 1,600 nodes, one GIN seed floods from step 18. We observed no such flood from the unmodified models on the other released sets we examined; their short trajectories limit the opportunity for a late flood. On 800-node ER graphs four hops deep, the degree input lowers our processor's node accuracy (\S\ref{sec:trigger}).

What removes the flood? Three changes remove every failure of our processor: clamping the degree input at inference, training without it, or adding 200 dense graphs to the training pool (\S\ref{sec:remedies}). With the carried state detached after each step, none of the 23 tested seeds floods, against 6 under \bptt{}. We call this rule \Kone{}; it is truncated backpropagation with a window of one step, and it shares initialization and data order with \bptt{} seed by seed. It does not protect the processor with sums or with a residual update, and two published remedies leave floods.

\paragraph{Contributions.} We show the premature flood on the benchmark's task and training distribution. Our six flooding GRU models fit every training hint; the benchmark models pass a separate reach-fit threshold. In our processor we locate it in the hidden state with a reset fixed in advance, and we test whether the state counts steps, which \citet{bansal2022overthinking} designed against but did not measure. We give a test that exposes both routes: deep graphs, dense ones among them, scored by how far from the source the search stays exact, with an audit of the range of every input and aggregate. We study one algorithm, BFS, because its hint is the reached set and its trajectory is as long as the graph is deep. Failure out of distribution after an exact fit is known in general \citep{yehudai2021local,damour2022underspecification}, and we make no general claim about recurrent extrapolation.

\section{Related work}
\label{sec:related}

\paragraph{Neural algorithmic reasoning.} CLRS \citep{velickovic2022clrs} and its sparse successor SALSA-CLRS \citep{minder2023salsaclrs} train message-passing processors with per-step hints on small graphs and test on larger ones. The SALSA-CLRS baselines, GIN, PGN and RecGNN \citep{xu2019gin,velickovic2020pgn,grotschla2022recgnn}, carry a latent state through all steps inside autograd and run, at test time, for the instance's ground-truth trajectory length. \citet{ibarz2022generalist} remove teacher forcing, feeding the model its own predicted hints, and \citet{rodionov2023nohints} train without intermediate supervision. \citet{mylonas2026insertion} show that the CLRS baseline for insertion sort optimizes its hints only weakly and that many of its intermediate representations can already be decoded to the sorted output before the reference execution ends, which they read as a shortcut to the final answer; our processor fits every hint and still declares nodes reached before the search arrives on graphs outside its training range.

Two designs are Markovian in the algorithm's state. Discrete NAR \citep{rodionov2025discrete} teacher-forces discrete ground-truth states so that no gradient crosses a step, and ForgetNet \citep{bohde2024markov} removes the historical embedding, arguing that historical state hurts out-of-distribution execution; our stateless block is close to ForgetNet (\S\ref{sec:remedies}), and \S\ref{sec:drift} gives that argument a mechanism. \citet{georgiev2024dear} solve for the output as a fixed point of the processor, and \citet{grotschla2022recgnn} and \citet{mathys2023floodecho} show recurrent and flood-and-echo networks extrapolating to paths of a thousand nodes or more. Table~\ref{tab:positioning} (Appendix~\ref{app:registrations}) compares these schemes with ours.

\paragraph{Deep thinking and test-time iteration.} \citet{schwarzschild2021algorithm} train recurrent networks on easy instances and run more iterations on hard ones. \citet{bansal2022overthinking} name \emph{overthinking} (degradation past the training iteration count) and prevent it with a recall connection and a progressive loss that restarts the recurrence from a random iteration and discards the gradients before it; none of their ablations separates the random start from the discarded gradients. \citet{anil2022path} promote path-independent solutions with random initial states; \citet{buitrago2025length} trace length-generalization failures of recurrent models to states never reached in training, and recurrent-depth models backpropagate through only their last steps: \citet{geiping2025recurrentdepth} through the last eight iterations, \citet{wang2025hrm} through a one-step approximation at the final updates. \citet{suleymanzade2026loopedflows} conjecture that a stop-gradient between steps makes shortcuts harder to learn; we test this on a different failure (\S\ref{sec:remedies}).

\paragraph{Size generalization.} When local structure such as degree changes with size, GNNs have ``bad'' global minima that do well on small graphs but fail on large graphs \citep{yehudai2021local}; the flood is such a minimum, in a recurrent reasoner's carried state. In CLRS, degree shift alone harms some algorithms, though not BFS \citep{mahdavi2023ood}; reasoners' latents show an ``inability to deal with values outside the range observed during training'' \citep{mirjanic2023latent}; and extrapolation relies on encoding the task's nonlinearities in the architecture or features \citep{xu2021extrapolate}, whereas our processor, given the BFS update, fails through an input BFS does not need. Hint-ReLIC enforces next-step invariance under trajectory-preserving augmentations \citep{bevilacqua2023causal}. Our clamps diagnose input sensitivity; we do not compare their efficacy with causal regularization. \citet{viakhirev2026shallow} relate an operator's regime (settling, marginal or drifting) to whether added iterations help.

\section{Setup}
\label{sec:setup}

We keep the benchmark's task, training distribution and evaluation, and add test graphs that are both deep and dense.

\paragraph{Task and protocol.} We use BFS from SALSA-CLRS \citep{minder2023salsaclrs} unchanged. Training uses 6,000 Erd\H{o}s--R\'enyi graphs drawn with the benchmark's own sampler and configuration, with its sizes $n\in\{4,7,11,13,16\}$ and its range of edge probabilities; we call them the pool. They follow the distribution of the benchmark's released training set of 10,000 graphs, on which the benchmark's own models train (\S\ref{sec:benchmark}). The six flooding GRU models reproduce every hint on this entire pool; other runs are checked through their logged training losses (Appendix~\ref{app:compute}). The test sets are our draw of the benchmark's ER and Delaunay families at $n\in\{16,80,160,800\}$, made with its sampler and configuration (Appendix~\ref{app:families}); we call a family at one size a \emph{cell}. The benchmark's Watts--Strogatz sets and its sets at $n{=}1{,}600$ serve only to check its own models (\S\ref{sec:benchmark}). Evaluation uses reference trajectory lengths; our processor runs to the longest in each batch of four (Appendix~\ref{app:families}). The metrics are the benchmark's node accuracy on the parent pointer $\pi$ and graph accuracy (all nodes correct). Our processor carries the benchmark's own hint, the binary \emph{reach mask} of reached nodes. We decode $\pi$ from the step at which each node is first declared reached: a node's parent is its lowest-index neighbor reached one step earlier, which is the benchmark's tie-break. This decode returns the benchmark's $\pi$ from correct timing on every test set. With incorrect timing, its lowest-index fallback can still choose the correct parent. We therefore score timing separately.

\paragraph{Families that need propagation over distance.} The benchmark's ER graphs have mean eccentricity 2.6 to 4.1 at $n$ from 16 to 800, so they do not test propagation over distance. Its Delaunay graphs reach a mean eccentricity of 17 at $n{=}800$. We add three families with exact BFS labels: degree-controlled ER graphs at $n{=}800$, drawn with the benchmark's sampler (mean degree 9 to 33), and two hand-constructed ones. Perfect mazes on $k\times k$ grids (maze$k$) have degree 2 and eccentricity 35 to 403. \emph{Blobs} (blob16x$l$) are chains of $l$ cliques of 16 nodes, with two random edges joining consecutive cliques; their mean eccentricity ranges from 11 at 8 cliques to 77 at 64 (108 at most). Blobs have the mean degree of the benchmark's $n{=}800$ ER graphs (15) and are deeper, so a failure that needs both density and depth can show on them.

\paragraph{The processor.} The GRU processor is one message-passing step with about 200,000 parameters. It computes a message with a SiLU activation \citep{elfwing2018silu},
\[ m_i = \mathrm{SiLU}\big(W[h_i,\overline{h}_{N(i)},e_i^x,e_i^f,\overline{e^f}_{N(i)},\max_{j\in N(i)}e_j^f]+b\big), \]
and then applies a GRU cell \citep{cho2014gru}, $h_i \leftarrow \mathrm{GRU}(m_i,h_i)$. The affine encoders $e^x$ and $e^f$ map the inputs to 128 dimensions (Appendix~\ref{app:families}). Here $x_i=[\mathbb{1}[i\text{ is the source}],\ \deg(i)/8]$, and $f_i\in\{0,1\}$ is the reach mask. The encoded neighborhood mean and coordinatewise max let one application express the BFS update (a node is reached if any neighbor was). A linear readout gives the logit that node $i$ is reached at this step. The inputs $x$ and the reach mask $f$ re-enter at every application; $f$ is the model's own thresholded prediction, kept monotone by a running maximum. The degree feature \xdeg$=\deg(i)/8$, a common node feature scaled by a constant, is part of our processor and not of the benchmark's baselines, which carry no such input. This processor permits controlled interventions on the state and input. Clamping it removes every failure of this processor (\S\ref{sec:trigger}); two variants without it, one with neighborhood sums and one with a residual update, still fail. The neighborhood terms $\overline{h}_{N(i)}$ and $\overline{e^f}_{N(i)}$ are means; the sum variant of \S\ref{sec:sums} replaces them by sums, drops the degree input and changes nothing else, and the residual variant replaces the GRU with $h\leftarrow\mathrm{LN}(h+m)$ to test a different state update.

\paragraph{Training.} Listing~\ref{lst:loop} is the training loop. Both rules use the binary cross-entropy against the hint at every step. Under \bptt{} the losses are summed and \texttt{backward} is called once. Under \Kone{}, \texttt{backward} is called at every step, and $h$ is detached. The hard reach threshold already blocks gradients through $f$, without a straight-through estimator. The parameter update is then the same sum of per-step gradients minus the terms in which step $t$'s loss reaches steps before $t$. Truncated backpropagation with window $K$ detaches every $K$ steps. Everything else is identical, including the hidden size of 128, the Adam optimizer \citep{kingma2015adam}, learning rate $2\cdot 10^{-3}$ with cosine decay, batch 16, 12,000 steps, gradient clipping at norm 1, and the random seed that fixes initialization and data order.

\begin{lstlisting}[caption={The training loop for one batch. \texttt{P} is the processor, \texttt{enc\_x} and \texttt{enc\_f} encode the inputs and the reach mask, \texttt{G} is the graph and \texttt{reach\_t} the hint at step $t$; $L$, the trajectory length, counts the initial step, so a rollout makes $L{-}1$ applications.},label={lst:loop}]
h, f = zeros(), source_mask()
loss = 0
for t in range(1, L):
    h = P(h, enc_x(x), enc_f(f), G)
    logit = readout(h)
    loss_t = BCE(logit, reach_t)
    next_f = (sigmoid(logit) > 0.5).float()
    f = maximum(f, next_f)
    if detach:  # the K=1 rule
        loss_t.backward()
        h = h.detach()
    else: loss = loss + loss_t
if not detach: loss.backward()
\end{lstlisting}

\paragraph{Statistics.} With no model at all, the decoder scores the floors of Appendix~\ref{app:floors}. The lowest-index and all-reached baselines have graph accuracy 0 on the tested sets with more than 16 nodes and at most 0.05 on the 16-node sets. Positive graph accuracy on the larger sets exceeds these two controls. Node accuracy does not show how far the search stays correct, so we also report the \emph{exact-timing radius}: the largest distance $d$ such that, at every distance up to $d$, at least 95\% of the nodes are declared reached at the right step. Nodes are pooled over the graphs of a test set. The 95\% cutoff is a summary tolerance, not a tie-breaking allowance; graph accuracy 1 means all parent pointers are correct.

The seed fixes initialization and data order, so seeds are paired across training choices: rates are compared with the exact McNemar test on the seeds whose outcomes differ, and node accuracies with exact sign tests. A rate observed as $0/23$ has a one-sided 95\% upper limit of 12\%. Intervals describe variation across seeds on fixed graph samples, not uncertainty over new graph draws. Appendix~\ref{app:registrations} gives the main comparisons with their intervals and Holm's correction (Table~\ref{tab:stats}), and what we expected from each experiment and what happened (Table~\ref{tab:registrations}). The correction is a descriptive summary of a retrospectively chosen family, not a confirmatory analysis.

\begin{table}[t]
\caption{Seeds (of 23) at graph accuracy 1 at both deep blob cells (blob) and at maze16 and maze32 (maze), and mean node accuracy on ER~800 and degree-17 ER, for each training choice under the pool. Progressive loss and random initial states $h_0$ are the literature's remedies added to \bptt{} (\S\ref{sec:remedies}); stateless is \Kone{} without a carried state; residual replaces the GRU cell; sum replaces the two neighborhood means. The ER columns use a second sample of 20 graphs per cell, drawn separately from those of Table~\ref{tab:headtohead}.}
\label{tab:rates}
\centering\footnotesize\setlength{\tabcolsep}{3pt}
\begin{tabular}{lrrrr}
\toprule
& \multicolumn{2}{c}{seeds perfect} & \multicolumn{2}{c}{node accuracy} \\
\cmidrule(lr){2-3}\cmidrule(lr){4-5}
rule & blob & maze & ER 800 & deg.\ 17 \\
\midrule
BPTT & 17/23 & 23/23 & 0.581 & 0.413 \\
truncated, $K{=}2$ & 16/23 & 23/23 & 0.587 & 0.425 \\
BPTT, progressive loss & 18/23 & 23/23 & 0.615 & 0.468 \\
BPTT, random $h_0$ & 22/23 & 23/23 & 0.538 & 0.346 \\
$K{=}1$ & 23/23 & 23/23 & 0.771 & 0.722 \\
stateless & 23/23 & 23/23 & 0.690 & 0.585 \\
BPTT, residual & 0/23 & 22/23 & 0.379 & 0.307 \\
$K{=}1$, residual & 0/23 & 17/23 & 0.653 & 0.536 \\
BPTT, no deg. & 23/23 & 23/23 & 1.000 & 1.000 \\
$K{=}1$, no deg. & 23/23 & 23/23 & 1.000 & 1.000 \\
BPTT, sum, no deg. & 8/23 & 23/23 & 0.899 & 0.893 \\
$K{=}1$, sum, no deg. & 13/23 & 23/23 & 0.812 & 0.772 \\
BPTT, residual, no deg. & 15/23 & 17/23 & 1.000 & 1.000 \\
$K{=}1$, residual, no deg. & 19/23 & 17/23 & 1.000 & 1.000 \\
\bottomrule
\end{tabular}
\end{table}

\begin{figure*}[t]
\centering\includegraphics[width=0.9\textwidth]{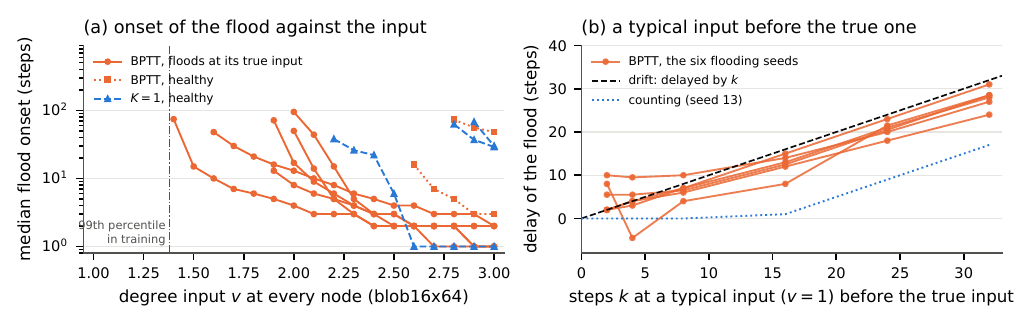}
\caption{The drift probe on blob16x64. (a) Median onset of the flood with the degree input set to one value $v$ at every node. The panel shows the six failing \bptt{} models, three healthy \bptt{} models (one does not flood up to 3.0) and three \Kone{} models. The dash-dotted line is the training pool's 99th percentile. (b) The switch: delay of the flood, relative to its onset $t^*$ under the true input, when the first $k$ steps run under a typical input ($v{=}1$), for the six failing models. A drifting state is delayed by $k$ (dashed line); a counting state is not delayed until $k$ reaches $t^*$ (dotted line, drawn for seed 13).}
\label{fig:drift}
\end{figure*}

\section{Results}
\label{sec:results}

\subsection{The premature flood}
\label{sec:flood}

At both deep blob cells, blob16x32 and blob16x64, \bptt{} reaches graph accuracy 1 in 17 of 23 seeds, truncated backpropagation with $K{=}2$ in 16 and \Kone{} in 23 (Table~\ref{tab:rates}, per seed Figure~\ref{fig:rates}; exact McNemar $p=0.031$ against \bptt{}, 0.16 after Holm's correction, Table~\ref{tab:stats}). We classify a model as \emph{healthy} if it is perfect at both deep cells, and as \emph{failing} otherwise. On mazes of eccentricity 403 all three rules are at 23/23, so this is not a general fragility of \bptt{} on deep graphs. A gradient across one step boundary is enough for failure at the deep cells: with $K{=}2$, 7 of 23 seeds fail, with the radius pattern of the \bptt{} floods (Appendix~\ref{app:truncation}).

The six failing \bptt{} models fail in the same way (Table~\ref{tab:failing}, Appendix~\ref{app:registrations}). No node is reached late, and 22 to 92\% are reached early. After a model-specific number of steps, the readout declares nodes reached early, so the exact-timing radius stops growing with the graph. Figure~\ref{fig:teaser}b shows one such model.

Nothing in the pool penalizes the flood. Of its graphs, 99\% are at most five hops deep and none is deeper than eight. After step five, only 156 of its 61,200 nodes are still being reached (Table~\ref{tab:pools}). On all 6,000 training graphs, each of the six models reproduces every hint at every step (Table~\ref{tab:density}).

\subsection{Where the flood is carried}
\label{sec:mechanism}

The weights do not change at inference, and the reach mask only accumulates the model's own decisions, so we test the hidden state $h$ by zeroing it at inference with the mask kept. A first test chose the period per seed after we had seen each onset; we then ran the test on all 46 models and ten test sets with the period fixed in advance and the same for every seed (Table~\ref{tab:reset}). Resetting every four steps takes \bptt{} from 17/23 to 22/23 perfect at both deep cells. It repairs five of the six floods to full radius (Figure~\ref{fig:flood}, Appendix~\ref{app:registrations}; the sixth floods by step 5). No model of either rule loses a hop of radius on any set.

\subsection{Counting or drift}
\label{sec:drift}

Does the state count steps, or does the input move it? If the state counts, it changes at every step whatever the input, and a tail input only makes the readout act on the count. If it drifts, it moves toward the flood only while the input is in the tail. We changed the degree input at inference on the saved weights of the six failing \bptt{} models and six healthy ones (three \bptt{}, three \Kone{}) on the 20 graphs of blob16x64 (Figure~\ref{fig:drift}; Tables~\ref{tab:drift} and \ref{tab:drift_onsets}). A graph's \emph{onset} is the first step at which a node is declared reached before the search arrives; $t^*$ is the onset under the true input.

Set to one value $v$ at every node, a higher input brings the flood earlier: seed 13 floods after 48 steps at $v{=}1.6$, 16 at 1.9, 4 at 2.5 and 2 at 3.0. Every model that floods follows the same fall (Figure~\ref{fig:drift}a; thresholds in \S\ref{sec:trigger}).

In the \emph{switch}, the input is held at a typical value, 1.0, for the first $k$ steps and then set to its true value. A counting state then floods at step $\max(t^*,k{+}1)$, and a drifting one at $k{+}t^*$. In the \emph{transplant}, a fresh rollout under the true input starts from a state that a rollout under the typical input had reached at step $k$. A counting state, having counted $k$ steps, then floods at $\max(1,t^*{-}k)$, and a drifting one, unmoved by the typical input, at $t^*$. We place each onset on a scale from drift's prediction (0) to counting's (1). By a rule fixed in advance, a model whose median is 0.75 or more counts, and one at 0.25 or less drifts. No model counts (Table~\ref{tab:drift}). Three drift and three lie between, and the transplant gives the same classes. In the three that lie between, time under the typical input brings the flood forward, more slowly than time under the true input. After 32 typical steps, seed 13 floods at step 40, where drift predicts 48 and counting 33 (Figure~\ref{fig:drift}b).

As a control, a state from a rollout under the true input brings the flood forward by $k$ in every failing model, as both readings predict; no healthy model floods under either intervention. By the rule we fixed in advance, the outcome is mixed. What the six share is that time under the typical input brings the flood forward by less than half of what counting predicts (Table~\ref{tab:drift}): the readout acts on where the state has moved, and a tail input moves it faster than a typical one.

\begin{figure}[t]
\centering\includegraphics[width=0.62\linewidth]{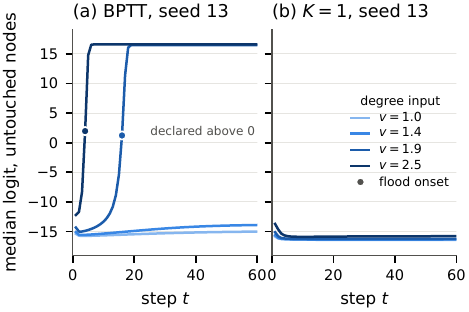}
\caption{The readout of untouched nodes (true distance greater than the step) on blob16x64, with the true reach mask fed in so that no declaration enters the input, under the degree input set to $v$ at every node. (a) The flooding \bptt{} model of seed 13, which first floods at $v{=}1.6$; dots mark the flood onsets of the model's own rollouts at each input. (b) The \Kone{} model of the same seed.}
\label{fig:drift_logit}
\end{figure}

At step $t$ a node more than $t$ hops from the source has received no influence from the reached set, which spreads one hop per step, so its state evolves as it would if the graph had no source at all. We ran the saved models with the true reach mask fed in at every step, so that no declaration, right or wrong, enters the input, and read the logit of these untouched nodes (Figure~\ref{fig:drift_logit}). Under the inputs at which a model floods (1.9 and 2.5 for the model of Figure~\ref{fig:drift_logit}a), their median logit climbs to the decision near the flood onset. Under the typical input it stays below. In each of the six failing models, the climb over steps 2 to 20 under the typical input is at most a fifth of the climb under $v{=}1.9$. The \Kone{} model of seed 13 stays flat under all four inputs (Figure~\ref{fig:drift_logit}b).

\subsection{What sets it off}
\label{sec:trigger}

What sets the flood off? On chains of 64 cliques of $q$ nodes (Table~\ref{tab:density}, Appendix~\ref{app:registrations}), the six failing models are exact to the last node for $q{=}4$ and $8$ (mean degree 4.0 and 7.5, 68 and 83 hops deep) and flood only at $q{=}16$, degree 15, so depth alone does not set it off. No maze floods under its true input.

Over the 61,200 nodes of the training pool, the degree feature \xdeg$=\deg/8$ has a mean of 0.62, a 99th percentile of 1.38 and a single node at its maximum, 1.88. Every blob node is at the training maximum or above. On the benchmark's ER $n{=}800$ cell, 71\% of nodes are at 1.5 or above, a value 0.4\% of training nodes reach. \label{sec:headtohead}Over 23 paired seeds, the two rules agree within 0.01 node accuracy on every ER and Delaunay cell except ER $n{=}800$ (Table~\ref{tab:headtohead}, Appendix~\ref{app:families}; at 160 nodes \Kone{} is already ahead in 18 of 21 untied seeds). On that cell and on our ER graphs of mean degree 17 at the same size, four hops deep, which together we call the band, the same tail input costs both rules node accuracy, \bptt{} more: on the head-to-head sample (Table~\ref{tab:headtohead}), \Kone{} leads by 0.15 on ER~800 and 0.22 on degree-17 graphs (in each cell \Kone{} is ahead in 19 seeds, behind in 3 and tied in 1; exact sign test $p<0.001$). On these four-hop graphs we did not separate early from late declarations; the clamp below repairs both cells with the floods.

We clamped the feature to 1.0 or 0.5 at inference on all 46 models of the GRU processor, as specified in advance (Table~\ref{tab:clamp}, Appendix~\ref{app:registrations}). Either clamp makes every model exact on all 21 test cells, the deep families together with new draws of the benchmark's cells (the combined sample, Appendix~\ref{app:families}), which repairs the six floods, the band and the high-degree cells. A processor trained without the feature is exact on every cell in every seed under either rule (Tables~\ref{tab:rates} and \ref{tab:archiveC}).

In the sweep of \S\ref{sec:drift}, the failing models first flood at a uniform input from 1.4 to 2.0, above the pool's 99th percentile and, for four of them, above its maximum. The six healthy models flood from 2.2 to 2.9, except one that never floods up to 3.0. A degree-2 maze never floods under its true input, but at an input of 1.9 it floods in all six failing models, with 0.90 to 0.99 of its nodes declared early within 150 steps, and in one healthy model, so the input's value alone sets the flood off.

With the feature, \bptt{} floods in 6 of 23 seeds, $K{=}2$ fails in 7 and \Kone{} in none. The three \Kone{} models we probed flood too, once the input reaches 2.2 or more, an outcome we had named as refuting \Kone{}'s protection. These seeds show different observed flood rates and input thresholds under the two training rules; the rate comparison does not survive Holm's correction.

\subsection{Sums}
\label{sec:sums}

Can an aggregate that grows with the graph break the processor without the degree input? A neighborhood sum, the aggregator of the original GIN \citep{xu2019gin}, grows with the number of neighbors whatever the features, and \citet{wittig2026which} note that unnormalized sums can void the conditions under which message-passing networks provably extrapolate. We replaced the processor's two neighborhood means by sums and removed the degree feature, changing nothing else. This breaks the reasoner at the deep cells in 15 of 23 seeds under \bptt{} and 10 of 23 under \Kone{}, where the means fail in none (exact McNemar $p<10^{-4}$ and $p=0.002$; Table~\ref{tab:rates}). Both rules fail; their rates are not separated at 23 seeds ($p=0.30$). The failures take two forms: 10 models flood, 14 stall, declaring nodes reached late or never, and one does both.

A blob node sums over 15 to 18 neighbors, against a median of 5 in the training pool. At inference we replaced a sum by $c$ times the neighborhood mean. Rescaling the hidden-state sum alone to $c{=}5$, the training median, makes 15 of the 25 failing models exact and breaks 1 of the 21 healthy ones (Table~\ref{tab:sum281}). The count is not the whole account: with both sums rescaled, 8 of the 25 still fail at the training median (the other probes: Appendix~\ref{app:registrations}).

\subsection{The benchmark's own models}
\label{sec:benchmark}

We trained SALSA-CLRS's own GIN, PGN and RecGNN with the benchmark's code, configurations, released training set and hints, ten seeds each (RecGNN nine), and the same GIN with its max aggregation replaced by a sum, and read their predicted reach hint at every step on our families and on the benchmark's own test sets (Tables~\ref{tab:salsa} and \ref{tab:p283}, Appendix~\ref{app:registrations}). The three unmodified models reproduce their published accuracy with hints. They take no degree input, aggregate with max and feed no prediction back, so our account of the first route predicted at most one flooding seed in each model. Among seeds whose early and late-or-never fractions are each below 1\% on the 16-node ER set, which we call fitting, GIN declares nodes reached early on blob16x64 in 3 of 7, PGN in 1 of 10 and RecGNN in none of 8. Each of the four flooding seeds does the same on the mazes, from nearly the same step as on the blobs (Table~\ref{tab:salsa_onsets}), and GIN seed 42 also on the first 200 graphs of the benchmark's Delaunay test set at 1,600 nodes, from step 18 or 19 in 199 graphs. One GIN seed stalls instead. Whether or not they flood, the three models' $\pi$ output fails on the deepest blobs (Table~\ref{tab:salsa}). GIN with sums declares nodes early in 7 of 7 from the first or second step (six floods, and one that also stalls), on the mazes too in 6 of 7. For the unmodified models this documents a nearly fixed onset across sparse and dense graphs, without identifying a state mechanism. These are reach-hint failures; their effect on the separately predicted $\pi$ was not measured on the seven released test sets.

\subsection{Remedies}
\label{sec:remedies}

What removes the flood? Three remedies remove every failure of the GRU processor: clamping the degree input, dropping it, and adding dense graphs to the pool. The clamp and the reset (\S\ref{sec:trigger}, \S\ref{sec:mechanism}) are interventions on trained models; each of the other remedies we compare is a separate training run, so each is an association between a training choice and an outcome (Tables~\ref{tab:rates}, \ref{tab:rates_pools} and \ref{tab:archiveC}).

\paragraph{Changing the input.} We extended the 6,000 training graphs with 2,000 hand-constructed graphs of one family at a time (Tables~\ref{tab:pools} and \ref{tab:rates_pools}, Figure~\ref{fig:horizon}). With a dense pool of 128- or 256-node blobs, \bptt{} floods in 0/23, against our prediction that a deeper pool would leave its rate unchanged, and both rules are perfect on every family, the band included. With a sparse pool of $4\times 24$ grids, \bptt{} still fails in 1/23 and the band stays open. A pool of 2,000 blob16x4 graphs, whose deepest is one hop shallower than the deepest training graph, puts every seed of both rules at graph accuracy 1 on all 21 cells (Table~\ref{tab:archiveC}), except one \bptt{} seed that misses one degree-33 graph. A pool of only 200 blob16x8 graphs does the same, and no seed misses a graph. A dense pool supplies degree inputs up to the blob cells' values, about 2.25, yet it also repairs the ER cells whose inputs reach 4.6, so coverage alone does not explain these repairs.

\paragraph{The residual variant.} With a layer-normalized residual update \citep{ba2016layernorm}, $h\leftarrow\mathrm{LN}(h+m)$, in place of the GRU cell, both rules fail at both deep cells in 23 of 23 seeds with low logged training loss, and the variant fails even without the degree input (Appendix~\ref{app:registrations}). Its \bptt{} floods begin at step two or three and survive resets every two or four steps, so they follow neither route (Table~\ref{tab:probe232}). A dense pool repairs it: with 2,000 blob16x4 or 200 blob16x8 graphs added to the pool, it is perfect at both deep cells in at least 22 of 23 seeds per rule.

\paragraph{Removing the carried state.} Zeroing $h$ before every application, in training and at inference, gives a stateless block close to ForgetNet \citep{bohde2024markov}. It is perfect at both deep cells and on the mazes in every seed, and on the band it lies between the two rules (Appendix~\ref{app:registrations}).

\paragraph{Removing the gradient through the carried state.} This is \Kone{}. No tested seed floods with the degree feature (0/23 against 6/23; Holm $p=0.16$), though its models flood once the input is raised further; it does not with sums (10/23 against 15/23, $p=0.30$) or for the residual variant (23/23 under both rules). We had said in advance that \Kone{}'s protection would not be general if \Kone{} failed at a rate not distinguishable from \bptt{}'s; that happened with the residual variant and with sums. \citet{suleymanzade2026loopedflows} conjecture that a stop-gradient between steps makes shortcuts harder to learn; the first case is consistent with that conjecture, while the other two show that detachment does not eliminate these failures.

\paragraph{Randomizing the start.} Two remedies from the literature change how training starts the recurrence, each added to \bptt{} over 23 seeds (Table~\ref{tab:remedies}, Appendix~\ref{app:registrations}). The progressive loss of \citet{bansal2022overthinking} leaves the flood in 5 of 23 seeds ($p=1$ against \bptt{}'s 6 and $p=0.06$ against \Kone{}'s 0), and random initial states \citep{anil2022path} leave it in 1 of 23, a rate that 23 seeds do not separate from either rule ($p=0.06$ and 1). Every failing seed under either remedy floods.

\section{Discussion}
\label{sec:discussion}

Why does the pool allow the flood? The hints fix the state's response for a few steps under inputs like the training ones. After step five almost every hint is the full mask, and 0.4\% of training nodes carry a degree input of 1.5 or more. A state that drifts toward \emph{reached} under an input the pool barely showed fits every hint, so the observed training hints do not distinguish it from one that propagates. Which solution a seed finds depends on training. With the degree input, 6 of 23 seeds under \bptt{} find the flooding one and none under \Kone{}; with sums, the size of the aggregate explains the failures only in part, and the benchmark's own models reach the flood by the second route. For the degree input this is underspecification in the sense of \citet{damour2022underspecification}, and here we can name its mechanism: a carried state that moves under an input in the tail of what training covered.

Two checks would expose these failures. An audit of the range of every input and aggregate under the size-extrapolation protocol catches the first route: a feature or sum growing with degree or size enters a thinly sampled part of its training distribution or goes beyond it, and a clamp or mean is a cheap control. Scoring graph accuracy and an exact-timing radius on deep graphs catches both routes, as long as some of them are dense. On the benchmark's shallow cells the search ends before a late flood can cost anything, and on our sparse test cells both rules stay near exact (Table~\ref{tab:headtohead}). The benchmark's unmodified models flood from step 18 or later (Table~\ref{tab:salsa_onsets}), so of its test sets only the Delaunay graphs at 1,600 nodes run past step 18 on every graph. Among the fitting seeds on the 200 graphs tested there, GIN seed 42 floods and no other seed does; on our draw of its Delaunay graphs at 800 nodes, no flooding seed declares a node early.

\section{Limitations}
\label{sec:limitations}

The flood appears in our processor and, by a second route, in the benchmark's own models, but we test one algorithm, BFS, and one family of our own processor. Of the other SALSA-CLRS tasks we ruled out two whose trajectories or updates cannot exercise the depth axis (Appendix~\ref{app:registrations}), and we did not test DFS, Dijkstra or eccentricity. Our processor comes with one training budget, so the rates are not claimed to be universal. For sums, the size of the aggregate does not explain every failure, we describe the stalls without explaining them, and the drift probe gives no single reading for their floods. The residual variant fails without a tail input, and its \bptt{} floods survive resets of $h$ every two or four steps; we do not explain them. Our account of the trigger holds for the GRU processor: the benchmark's own models flood from nearly the same step on blobs and mazes, and we did not probe what their state carries. The onset criterion does not rule out every counting component. It probes six models with a synthetic uniform input: we did not identify what inside $h$ moves, and three of the six lie between drift and counting. The literature's remedies were each tested at one setting and only under \bptt{}. Over the main comparisons, Holm's correction keeps only the comparisons of sums against means and of the two rules on ER $n{=}800$; the rule comparison at the deep cells ($p=0.031$, 0.16 corrected) and the rest do not survive it (Table~\ref{tab:stats}).

\section{Conclusion}
\label{sec:conclusion}

A recurrent reasoner that fits every hint of its training graphs can still flood deep graphs, declaring nodes reached before the search arrives. In our GRU processor, a degree input in the tail of the training distribution moves the hidden state toward the flood; without that input, summing the neighbors' states instead of averaging them also breaks the reasoner, by floods and by stalls. In two of the benchmark's own models, some seeds flood from a fixed step, one of them on the benchmark's own deepest test set. Clamping or dropping the input, or adding 200 dense training graphs, removes every failure of the GRU processor. Benchmarks that extrapolate in size should audit the range of every input and aggregate their reasoners read, and test on deep graphs, dense ones among them.

\subsection*{AI Use Statement}

AI assistance was used in writing the manuscript and the code, and in preparing the figures. The authors designed the study, ran and checked the experiments, and take full responsibility for the content.

\renewcommand{\bibsection}{\subsubsection*{References}}
\bibliographystyle{plainnat}
\bibliography{refs}

\clearpage
\appendix
\renewcommand{\floatpagefraction}{0.75}\renewcommand{\dblfloatpagefraction}{0.75}
\thispagestyle{empty}
\begin{center}{\Large\bfseries Appendix}\end{center}
\section{Model-free reference points}
\label{app:floors}

What does the decoder score with no model at all? Before any model number was read, the decoder of \S\ref{sec:setup} was scored on every test set under three rules: a perfect reach mask, which must decode to the benchmark's $\pi$ or the decoder is wrong; the lowest-index rule, in which every node names its lowest-index neighbor, which is what the decoder falls back to when no neighbor was reached one step earlier; and the degenerate model that declares every node reached at step~1. The perfect reach mask decodes to $\pi$ at node and graph accuracy 1 on every set of both samples. Table~\ref{tab:floors} gives the other two.

\begin{table}[!htbp]
\caption{Model-free reference points, node/graph accuracy. (A): the benchmark sample; (B): the deep sample (Appendix~\ref{app:floors}). A perfect reach mask scores 1/1 everywhere and is omitted. n/a: the all-reached rule was not scored on the benchmark sample's degree-controlled cells.}
\label{tab:floors}
\centering\footnotesize\setlength{\tabcolsep}{3pt}
\begin{tabular}{lrrrr}
\toprule
& \multicolumn{2}{c}{lowest-index rule} & \multicolumn{2}{c}{all reached at 1} \\
\cmidrule(lr){2-3}\cmidrule(lr){4-5}
test set & node & graph & node & graph \\
\midrule
ER 16 (A) & 0.327 & 0.000 & 0.613 & 0.025 \\
ER 80 (A) & 0.285 & 0.000 & 0.380 & 0.000 \\
ER 160 (A) & 0.249 & 0.000 & 0.311 & 0.000 \\
ER 800 (A) & 0.240 & 0.000 & 0.257 & 0.000 \\
Delaunay 16 (A) & 0.257 & 0.000 & 0.523 & 0.025 \\
Delaunay 80 (A) & 0.270 & 0.000 & 0.329 & 0.000 \\
Delaunay 160 (A) & 0.269 & 0.000 & 0.301 & 0.000 \\
Delaunay 800 (A) & 0.264 & 0.000 & 0.270 & 0.000 \\
ER 800, deg.\ 9 (A) & 0.278 & 0.000 & n/a & n/a \\
ER 800, deg.\ 13 (A) & 0.236 & 0.000 & n/a & n/a \\
ER 800, deg.\ 17 (A) & 0.222 & 0.000 & n/a & n/a \\
ER 800, deg.\ 25 (A) & 0.246 & 0.000 & n/a & n/a \\
ER 800, deg.\ 33 (A) & 0.221 & 0.000 & n/a & n/a \\
ER 800 (B) & 0.226 & 0.000 & 0.242 & 0.000 \\
ER 800, deg.\ 17 (B) & 0.216 & 0.000 & 0.236 & 0.000 \\
blob16x8 (B) & 0.147 & 0.000 & 0.254 & 0.000 \\
blob16x16 (B) & 0.145 & 0.000 & 0.199 & 0.000 \\
blob16x32 (B) & 0.175 & 0.000 & 0.203 & 0.000 \\
blob16x64 (B) & 0.185 & 0.000 & 0.199 & 0.000 \\
maze8 (B) & 0.540 & 0.000 & 0.556 & 0.000 \\
maze16 (B) & 0.539 & 0.000 & 0.543 & 0.000 \\
maze24 (B) & 0.528 & 0.000 & 0.529 & 0.000 \\
maze32 (B) & 0.537 & 0.000 & 0.538 & 0.000 \\
\bottomrule
\end{tabular}
\end{table}

Node accuracy on this task has a floor of 0.14 to 0.33 under the lowest-index rule on the ER, Delaunay and blob families, higher under the all-reached rule on the smallest sets, up to 0.61 at $n{=}16$. On mazes the floor is 0.53 to 0.56, because degree~2 makes the lowest-index neighbor the true parent about half the time by accident. A node-accuracy claim on a maze would be a claim about the decoder, which is why the paper reports graph accuracy and the exact-timing radius there. The lowest-index and all-reached baselines have graph accuracy 0 on every tested set with more than 16 nodes, and at most 0.025 on the 16-node sets of the first two samples (0.05 on the combined sample's ER~16 set). Positive graph accuracy exceeds these two baselines; it is not a guarantee of correct hint timing.

The \emph{benchmark sample} is our draw of the benchmark's ER and Delaunay test cells, made with its sampler and configuration, together with the degree-controlled family (Table~\ref{tab:headtohead}). Its original graph archive is not included in the supplement, so its model-free floors are preserved as archived values. The deep and combined samples are included and their floors are recomputed. The \emph{deep sample} holds the blobs and mazes with their own ER anchors (Tables~\ref{tab:rates}, \ref{tab:failing}, \ref{tab:reset}). The \emph{combined sample} of \S\ref{sec:trigger} is the deep sample plus new draws of the benchmark sample's cells, whose floors were computed the same way before any model number: lowest-index node accuracy 0.23 to 0.30, graph 0, except 0.05 for the all-reached rule on its ER~16 set.

\FloatBarrier
\section{Families and training pools}
\label{app:families}

Which graphs do the models train and test on?

\begin{table*}[!htbp]
\caption{Head-to-head over 23 paired seeds on the benchmark sample: our draw of the benchmark's ER and Delaunay test cells, made with its sampler and configuration, and the degree-controlled ER family. Node accuracy, graph accuracy and the exact-timing radius are means over seeds. ``Sign'' counts seeds where \Kone{} exceeds, trails and ties \bptt{}. At degree 33 both rules are near the model-free floor; at degree 25 \bptt{} is near it and \Kone{} above it (Appendix~\ref{app:floors}). Tables~\ref{tab:rates} and \ref{tab:clamp} use a second sample of the ER~800 and degree-17 cells.}
\label{tab:headtohead}
\centering\footnotesize\setlength{\tabcolsep}{3.5pt}
\begin{tabular}{lrrrrrrrrrrr}
\toprule
& & & \multicolumn{4}{c}{node accuracy on $\pi$} & \multicolumn{2}{c}{graph accuracy} & \multicolumn{3}{c}{exact-timing radius (hops)} \\
\cmidrule(lr){4-7}\cmidrule(lr){8-9}\cmidrule(lr){10-12}
test set & ecc & deg & BPTT & K$=$1 & paired $\Delta$ & sign & BPTT & K$=$1 & BPTT & K$=$1 & sign \\
\midrule
ER 16 & 2.6 & 6.4 & 1.000 & 1.000 & +0.000 & 0+/0$-$/23= & 1.000 & 1.000 & 4 & 4 & 0+/0$-$/23= \\
ER 80 & 3.5 & 9.4 & 0.998 & 0.999 & +0.000 & 3+/2$-$/18= & 0.991 & 0.992 & 7 & 7 & 0+/0$-$/23= \\
ER 160 & 3.6 & 11.6 & 0.972 & 0.979 & +0.007 & 18+/3$-$/2= & 0.867 & 0.914 & 4.74 & 4.61 & 1+/2$-$/20= \\
ER 800 & 4.1 & 15.1 & 0.651 & 0.802 & +0.151 & 19+/3$-$/1= & 0.378 & 0.567 & 1.43 & 2.91 & 12+/0$-$/11= \\
Delaunay 16 & 3.3 & 5.7 & 1.000 & 1.000 & +0.000 & 0+/0$-$/23= & 1.000 & 1.000 & 4 & 4 & 0+/0$-$/23= \\
Delaunay 80 & 6.8 & 6.6 & 1.000 & 1.000 & +0.000 & 0+/0$-$/23= & 1.000 & 1.000 & 8 & 8 & 0+/0$-$/23= \\
Delaunay 160 & 8.9 & 6.8 & 1.000 & 1.000 & +0.000 & 0+/0$-$/23= & 1.000 & 1.000 & 10 & 10 & 0+/0$-$/23= \\
Delaunay 800 & 16.6 & 6.9 & 1.000 & 1.000 & +0.000 & 0+/0$-$/23= & 1.000 & 1.000 & 18 & 18 & 0+/0$-$/23= \\
ER 800, deg.\ 9 & 4.9 & 9.0 & 0.999 & 0.999 & +0.000 & 1+/1$-$/21= & 0.996 & 0.996 & 5 & 5 & 0+/0$-$/23= \\
ER 800, deg.\ 13 & 4.0 & 13.0 & 0.884 & 0.911 & +0.026 & 9+/3$-$/11= & 0.540 & 0.728 & 3.17 & 3.48 & 7+/1$-$/15= \\
ER 800, deg.\ 17 & 3.9 & 16.9 & 0.532 & 0.751 & +0.219 & 19+/3$-$/1= & 0.011 & 0.377 & 1.57 & 2.57 & 15+/1$-$/7= \\
ER 800, deg.\ 25 & 3.0 & 24.9 & 0.261 & 0.380 & +0.119 & 14+/8$-$/1= & 0.000 & 0.051 & 1 & 1.35 & 6+/0$-$/17= \\
ER 800, deg.\ 33 & 3.0 & 33.0 & 0.261 & 0.261 & +0.000 & 7+/14$-$/2= & 0.000 & 0.000 & 1 & 1 & 0+/0$-$/23= \\
\bottomrule
\end{tabular}
\end{table*}
\paragraph{Implementation details.} The input encoder is affine $2\to128$, the reach encoder affine $1\to128$, and the message map affine $768\to128$, followed by SiLU. The GRU has input and hidden size 128; the readout is affine $128\to1$. The module registers 198,529 parameters, including 256 layer-normalization parameters used only by the residual variant; 198,273 participate in the GRU forward pass. Listing~\ref{lst:loop} uses elementwise \texttt{maximum}; $h$ starts at zero and $f$ marks only the source. The code also detaches $f$ under $K{=}1$, which is redundant because thresholding has already stopped its gradient. A training batch runs to its longest trajectory. Shorter graphs keep receiving their final reach target and contribute loss at every remaining step; no temporal mask is used. The standard evaluation and saved-weight range probe also run to the longest trajectory in each batch of four, without a per-graph stop mask. Shorter graphs can therefore receive extra steps for late predictions, but those steps cannot create an early declaration. The benchmark models' evaluator instead masks each graph at its own horizon. Dividing degree by eight is a fixed rescaling, not clipping or a training-range bound.

Continuous means in the head-to-head, extended pools, combined-sample and clamp tables have sample standard deviations across seeds (denominator $n-1$) in the supplementary analysis's \texttt{numbers.json}, under keys ending in \texttt{sd}. The benchmark parent-accuracy deviations condition on the same fitting seeds as its means. Table~\ref{tab:dispersion} gives the head-to-head deviations. These summaries quantify seed variation on each fixed graph draw, not between-pool uncertainty.

\paragraph{The pool and the benchmark's cells.} The pool and our ER and Delaunay test cells are drawn with the SALSA-CLRS sampler and its own configuration: training on Erd\H{o}s--R\'enyi graphs with $n\in\{4,7,11,13,16\}$ and edge probability drawn from $[\log 16/16,\ 3\log 16/16]$ at every training size, as the released code sets it, which gives an expected 2.6 to 7.8 neighbors per node at $n{=}16$ and a mean degree of 4.5 over the pool (each graph's mean degree, self-loop included, averaged over graphs); test ER graphs at $n$ with probability drawn from $[\log n/n,\ 3\log n/n]$; Delaunay graphs at $n$. The head-to-head, truncation and learning-rate runs (Tables~\ref{tab:headtohead} and \ref{tab:trunc}) use a separate draw of the pool and of the ER and Delaunay cells, made with the same sampler and configuration, and the two pools differ; comparisons across these branches are not matched pool effects. Every other run of our processor trains on the draw described here and in Table~\ref{tab:pools}, except the grids-and-paths runs of Appendix~\ref{app:registrations}, which train on grids. The hint is the reach mask, which we verified equals $[\mathrm{dist}\le t]$ on every trajectory, and the trajectory length is the number of hint steps. For context, the benchmark's published GIN, PGN and RecGNN baselines trained without hints report node accuracy 98 to 99\% and graph accuracy 2.2, 0.2 and 4.6\% on ER $n{=}800$, and on Delaunay $n{=}800$ node accuracy 53 to 67\% with graph accuracy 0\%. The representations differ, so neither direction is a matched comparison with our processor. The benchmark's own models of \S\ref{sec:benchmark} are trained with hints and checked against the accuracy published for its models trained with hints (Table~\ref{tab:salsa}).

\paragraph{The head-to-head.} Do the two rules differ on the benchmark's cells? Over 23 paired seeds (Table~\ref{tab:headtohead}), both are above 0.97 node accuracy and within 0.01 of each other on ER $n\le 160$ and on every Delaunay cell, and exact on Delaunay $n{=}800$, 17 hops deep on average. Both fail ER $n{=}800$ (node accuracy 0.65 under \bptt{} and 0.80 under \Kone{}). They differ most on the band, the 800-node ER graphs of mean degree 15 to 17, four hops deep, where \Kone{} leads by 0.15 on ER~800 and 0.22 on degree-17 graphs node accuracy ($19{+}/3{-}/1{=}$ in each cell, $p<0.001$); at degree 25 and 33 both fail, \bptt{} near the model-free floor. These failures and the flood of \S\ref{sec:flood} share one cause, the tail of the degree input (\S\ref{sec:trigger}).

\paragraph{Added families.} \emph{Degree-controlled ER:} the benchmark's own sampler at $n{=}800$ with a fixed edge probability $p=\bar d/(n-1)$ for target degree $\bar d\in\{8,12,16,24,32\}$, all above the connectivity threshold. Eccentricity falls from 4.9 to 3.0 as degree rises, so the family isolates degree with depth moving the other way. The sampler adds a self-loop to every node, which the degree feature and the quoted degrees of every sampled cell include (observed 9.0, 13.0, 16.9, 24.9 and 33.0 for these targets); the hand-constructed families have none. This shifts the degree feature by $1/8$ relative to an otherwise identical graph with self-loops. The feature-clamp experiments intervene on its value directly; the self-loop control for sums is reported separately in Appendix~\ref{app:registrations}.

\emph{Mazes} (maze$k$): perfect mazes on $k\times k$ grids by randomized depth-first search, a spanning tree with dead ends and mean degree~2, $k\in\{8,16,24,32\}$. \emph{Blobs} (blob16x$l$): chains of $l$ cliques of $q{=}16$ nodes (the generator's internal edge probability is $15/15$, so every block is complete), consecutive cliques joined by two random edges, connectivity checked and the graph resampled otherwise; $l\in\{4,8,16,32,64\}$. \emph{Grids} (grid4x$l$): $w\times l$ grids with 4-neighbor edges, $w{=}4$. In these hand-constructed families every label is produced by a hand-written BFS on the constructed graph, with the parent of a node its lowest-index neighbor at distance one less, which is the tie-break of the benchmark's decoder, and the BFS source is a node drawn uniformly at random by the family's generator. Generator seeds are fixed per family and disjoint from every test set's, so no training graph is a test graph.

Table~\ref{tab:families} lists the test sets with the properties the decoder reads from them. Table~\ref{tab:pools} lists the training pools of \S\ref{sec:remedies}, generated inside the training process from fixed seeds and checked graph for graph against the generators of the test sets. For the experiments of \S\ref{sec:trigger} and for checking that the pool remedy costs nothing on the benchmark's cells, the deep sample was combined with new draws of the benchmark sample's cells (ER 16/80/160 with 40 graphs, Delaunay 16/80/160/800 with 40/40/40/20, degree-controlled ER at 9, 13, 25 and 33 with 20), produced by the benchmark's sampler with the same seed formula. The sampler is not bit-reproducible across environments, so these are new draws and are never presented as the benchmark sample's graphs. This combined sample has 21 cells, and Table~\ref{tab:archiveC} gives all of them under the pool and under each remedy.

\begin{table}[!htbp]
\caption{Test sets. The benchmark sample above the rule, the deep sample below; ecc., deg.\ and $L$ are the mean eccentricity of the BFS source, the mean degree and the mean trajectory length.}
\label{tab:families}
\centering\footnotesize\setlength{\tabcolsep}{3pt}
\begin{tabular}{lrrrrr}
\toprule
test set & $n$ & graphs & ecc.\ & deg.\ & $L$ \\
\midrule
ER 16 & 16 & 40 & 2.6 & 6.4 & 3.6 \\
ER 80 & 80 & 40 & 3.5 & 9.4 & 4.5 \\
ER 160 & 160 & 40 & 3.6 & 11.6 & 4.6 \\
ER 800 & 800 & 20 & 4.1 & 15.1 & 5.1 \\
Delaunay 16 & 16 & 40 & 3.3 & 5.7 & 4.3 \\
Delaunay 80 & 80 & 40 & 6.8 & 6.6 & 7.8 \\
Delaunay 160 & 160 & 40 & 8.9 & 6.8 & 9.9 \\
Delaunay 800 & 800 & 20 & 16.6 & 6.9 & 17.6 \\
ER 800, deg.\ 9 & 800 & 12 & 4.9 & 9.0 & 5.9 \\
ER 800, deg.\ 13 & 800 & 12 & 4.0 & 13.0 & 5.0 \\
ER 800, deg.\ 17 & 800 & 12 & 3.9 & 16.9 & 4.9 \\
ER 800, deg.\ 25 & 800 & 12 & 3.0 & 24.9 & 4.0 \\
ER 800, deg.\ 33 & 800 & 12 & 3.0 & 33.0 & 4.0 \\
\midrule
ER 800 (B) & 800 & 20 & 4.1 & 15.1 & 5.1 \\
ER 800, deg.\ 17 (B) & 800 & 20 & 4.0 & 17.0 & 5.0 \\
blob16x8 & 128 & 40 & 10.6 & 15.2 & 11.6 \\
blob16x16 & 256 & 40 & 20.5 & 15.2 & 21.5 \\
blob16x32 & 512 & 20 & 40.6 & 15.2 & 41.6 \\
blob16x64 & 1,024 & 20 & 76.6 & 15.2 & 77.6 \\
maze8 & 64 & 40 & 35.4 & 2.0 & 36.4 \\
maze16 & 256 & 40 & 111.5 & 2.0 & 112.5 \\
maze24 & 576 & 20 & 229.3 & 2.0 & 230.3 \\
maze32 & 1,024 & 20 & 402.9 & 2.0 & 403.9 \\
\bottomrule
\end{tabular}
\end{table}

\begin{figure}[!htbp]
\centering\includegraphics[width=0.62\textwidth]{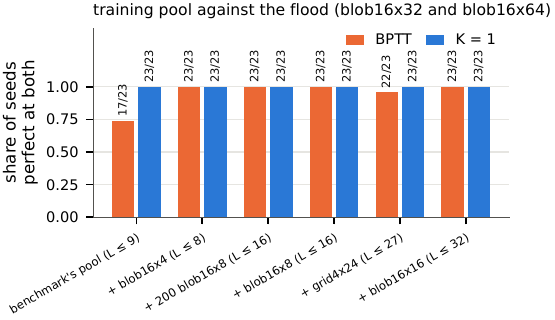}
\caption{The training-pool axis: seeds perfect at both deep blob cells, 23 seeds per pool. Each pool adds 2,000 graphs of one family to the 6,000 training graphs unless marked 200. Every dense pool, including one no deeper than the training graphs and one of 200 graphs, removes the flood for \bptt{}; Table~\ref{tab:rates_pools} gives the band, which the dense pools also close and the sparse grid pool does not.}
\label{fig:horizon}
\end{figure}

\begin{table*}[!htbp]
\caption{Training pools. Each extended pool is the 6,000 training graphs plus the graphs listed. ``Mean deg.'' is each graph's mean degree averaged over graphs; in the first row it counts the self-loop the benchmark's sampler adds to every node, which the hand-constructed graphs of the other rows do not have. The degree feature's mean in \S\ref{sec:trigger} is taken over nodes, which weights the larger and denser training graphs more and gives a higher value. The last two columns count nodes (over the whole pool) whose true BFS distance is at least 6 and at least 9, that is, nodes still being reached at late steps.}
\label{tab:pools}
\centering\footnotesize\setlength{\tabcolsep}{4pt}
\begin{tabular}{lrrrrrrrr}
\toprule
& & & & \multicolumn{2}{c}{eccentricity} & & \multicolumn{2}{c}{nodes first reached at step} \\
\cmidrule(lr){5-6}\cmidrule(lr){8-9}
pool & graphs & mean $n$ & mean deg.\ & mean & max & $P(\mathrm{ecc}\ge 5)$ & $\ge 6$ & $\ge 9$ \\
\midrule
SALSA-CLRS (benchmark) & 6,000 & 10 & 4.5 & 2.7 & 8 & 0.05 & 156 & 0 \\
blob16x4 $\times$ 2,000 & 2,000 & 64 & 15.2 & 5.5 & 7 & 0.83 & 13,789 & 0 \\
blob16x8 $\times$ 200 & 200 & 128 & 15.2 & 10.9 & 15 & 1.00 & 11,687 & 5,751 \\
blob16x8 $\times$ 2,000 & 2,000 & 128 & 15.2 & 11.0 & 15 & 1.00 & 115,789 & 58,245 \\
grid4x24 $\times$ 2,000 & 2,000 & 96 & 3.4 & 20.0 & 26 & 1.00 & 130,414 & 93,915 \\
blob16x16 $\times$ 2,000 & 2,000 & 256 & 15.2 & 21.6 & 31 & 1.00 & 357,885 & 277,776 \\
\bottomrule
\end{tabular}
\end{table*}

\begin{table*}[!htbp]
\caption{Rates under the extended training pools, 23 seeds per row: the 6,000 training graphs plus 2,000 (or 200) hand-constructed graphs of one family. Columns as in Table~\ref{tab:rates}; every row is a separate training run.}
\label{tab:rates_pools}
\centering\footnotesize\setlength{\tabcolsep}{3.5pt}
\begin{tabular}{llrrrr}
\toprule
& & \multicolumn{2}{c}{seeds perfect} & \multicolumn{2}{c}{node accuracy} \\
\cmidrule(lr){3-4}\cmidrule(lr){5-6}
rule & training pool & blob & maze & ER 800 & deg.\ 17 \\
\midrule
BPTT & +2,000 blob16x4 ($L\le 8$) & 23/23 & 23/23 & 1.000 & 1.000 \\
$K{=}1$ & +2,000 blob16x4 ($L\le 8$) & 23/23 & 23/23 & 1.000 & 1.000 \\
BPTT & +200 blob16x8 ($L\le 16$) & 23/23 & 23/23 & 1.000 & 1.000 \\
$K{=}1$ & +200 blob16x8 ($L\le 16$) & 23/23 & 23/23 & 1.000 & 1.000 \\
BPTT & +2,000 blob16x8 ($L\le 16$) & 23/23 & 23/23 & 1.000 & 1.000 \\
$K{=}1$ & +2,000 blob16x8 ($L\le 16$) & 23/23 & 23/23 & 1.000 & 1.000 \\
BPTT & +2,000 grid4x24 ($L\le 27$) & 22/23 & 23/23 & 0.441 & 0.247 \\
$K{=}1$ & +2,000 grid4x24 ($L\le 27$) & 23/23 & 23/23 & 0.521 & 0.354 \\
BPTT & +2,000 blob16x16 ($L\le 32$) & 23/23 & 23/23 & 1.000 & 1.000 \\
$K{=}1$ & +2,000 blob16x16 ($L\le 32$) & 23/23 & 23/23 & 1.000 & 1.000 \\
BPTT, residual & +2,000 blob16x4 ($L\le 8$) & 23/23 & 21/23 & 1.000 & 1.000 \\
$K{=}1$, residual & +2,000 blob16x4 ($L\le 8$) & 23/23 & 23/23 & 1.000 & 1.000 \\
BPTT, residual & +200 blob16x8 ($L\le 16$) & 22/23 & 23/23 & 1.000 & 1.000 \\
$K{=}1$, residual & +200 blob16x8 ($L\le 16$) & 23/23 & 23/23 & 1.000 & 1.000 \\
\bottomrule
\end{tabular}
\end{table*}

\begin{table*}[!htbp]
\caption{The combined sample's 21 cells, node/graph accuracy averaged over 23 seeds per rule: the GRU processor under the pool (both rules), then the minimum over the two rules under each remedy. A dagger marks a cell where the two rules differ by at least 0.005. Separately, the residual variant has seed-level imperfections on the mazes under the blob16x4 pool, and on blob16x64 and degree 33 under the 200-graph pool; every seed remains above 0.95 node accuracy. Under the original pool it fails Delaunay 800 (BPTT 0.44, $K{=}1$ 0.78), not shown here.}
\label{tab:archiveC}
\centering\footnotesize\setlength{\tabcolsep}{2.5pt}
\begin{tabular}{lrrrrrrr}
\toprule
& \multicolumn{2}{c}{benchmark pool (GRU)} & \multicolumn{3}{c}{GRU processor, min of two rules} & \multicolumn{2}{c}{residual variant, min of two rules} \\
\cmidrule(lr){2-3}\cmidrule(lr){4-6}\cmidrule(lr){7-8}
cell (node/graph) & BPTT & $K{=}1$ & $+$blob16x4 & $+$200 blob16x8 & no deg. & $+$blob16x4 & $+$200 blob16x8 \\
\midrule
ER 16 & 1.000/1.000 & 1.000/1.000 & 1.000/1.000 & 1.000/1.000 & 1.000/1.000 & 1.000/1.000 & 1.000/1.000 \\
ER 80 & 0.999/0.993 & 0.999/0.989 & 1.000/1.000 & 1.000/1.000 & 1.000/1.000 & 1.000/1.000 & 1.000/1.000 \\
ER 160 & 0.969/0.824 & 0.981/0.917 & 1.000/1.000 & 1.000/1.000 & 1.000/1.000 & 1.000/1.000 & 1.000/1.000 \\
ER 800 & 0.581/0.226 & 0.771/0.520 & 1.000/1.000 & 1.000/1.000 & 1.000/1.000 & 1.000/1.000 & 1.000/1.000 \\
Delaunay 16 & 1.000/1.000 & 1.000/1.000 & 1.000/1.000 & 1.000/1.000 & 1.000/1.000 & 1.000/1.000 & 1.000/1.000 \\
Delaunay 80 & 1.000/1.000 & 1.000/1.000 & 1.000/1.000 & 1.000/1.000 & 1.000/1.000 & 1.000/1.000 & 1.000/1.000 \\
Delaunay 160 & 1.000/1.000 & 1.000/1.000 & 1.000/1.000 & 1.000/1.000 & 1.000/1.000 & 1.000/1.000 & 1.000/1.000 \\
Delaunay 800 & 1.000/1.000 & 1.000/1.000 & 1.000/1.000 & 1.000/1.000 & 1.000/1.000 & 1.000/1.000 & 1.000/1.000 \\
ER 800, deg.\ 9 & 0.998/0.980 & 0.998/0.989 & 1.000/1.000 & 1.000/1.000 & 1.000/1.000 & 1.000/1.000 & 1.000/1.000 \\
ER 800, deg.\ 13 & 0.854/0.398 & 0.924/0.770 & 1.000/1.000 & 1.000/1.000 & 1.000/1.000 & 1.000/1.000 & 1.000/1.000 \\
ER 800, deg.\ 17 & 0.413/0.011 & 0.722/0.363 & 1.000/1.000 & 1.000/1.000 & 1.000/1.000 & 1.000/1.000 & 1.000/1.000 \\
ER 800, deg.\ 25 & 0.276/0.000 & 0.335/0.002 & 1.000/1.000 & 1.000/1.000 & 1.000/1.000 & 1.000/1.000 & 1.000/1.000 \\
ER 800, deg.\ 33 & 0.270/0.000 & 0.264/0.000 & 1.000/0.998 & 1.000/1.000 & 1.000/1.000 & 1.000/1.000 & 0.993/0.954$^\dagger$ \\
\midrule
blob16x8 & 0.985/0.955 & 1.000/1.000 & 1.000/1.000 & 1.000/1.000 & 1.000/1.000 & 1.000/1.000 & 1.000/1.000 \\
blob16x16 & 0.966/0.886 & 1.000/1.000 & 1.000/1.000 & 1.000/1.000 & 1.000/1.000 & 1.000/1.000 & 1.000/1.000 \\
blob16x32 & 0.935/0.796 & 1.000/1.000 & 1.000/1.000 & 1.000/1.000 & 1.000/1.000 & 1.000/1.000 & 1.000/1.000 \\
blob16x64 & 0.903/0.748 & 1.000/1.000 & 1.000/1.000 & 1.000/1.000 & 1.000/1.000 & 1.000/1.000 & 0.999/0.987$^\dagger$ \\
maze8 & 1.000/1.000 & 1.000/1.000 & 1.000/1.000 & 1.000/1.000 & 1.000/1.000 & 0.992/0.938$^\dagger$ & 1.000/1.000 \\
maze16 & 1.000/1.000 & 1.000/1.000 & 1.000/1.000 & 1.000/1.000 & 1.000/1.000 & 0.970/0.913$^\dagger$ & 1.000/1.000 \\
maze24 & 1.000/1.000 & 1.000/1.000 & 1.000/1.000 & 1.000/1.000 & 1.000/1.000 & 0.963/0.913$^\dagger$ & 1.000/1.000 \\
maze32 & 1.000/1.000 & 1.000/1.000 & 1.000/1.000 & 1.000/1.000 & 1.000/1.000 & 0.961/0.913$^\dagger$ & 1.000/1.000 \\
\bottomrule
\end{tabular}
\end{table*}

\FloatBarrier
\section{What we expected, and what happened}
\label{app:registrations}

\begin{table}[!htbp]
\caption{Training schemes for iterated graph reasoners: what is carried across an application boundary, and whether the loss gradient crosses it. Truncated BPTT backpropagates through the last $K$ applications; DEAR is the equilibrium reasoner of \citet{georgiev2024dear} and DNAR the discrete reasoner of \citet{rodionov2025discrete}.}
\label{tab:positioning}
\centering\footnotesize\setlength{\tabcolsep}{3pt}
\begin{tabular}{lll}
\toprule
scheme & carried & gradient \\
\midrule
SALSA baselines & own state & full \\
truncated BPTT & own state & last $K$ \\
DEAR & fixed point & implicit \\
DNAR & true hints & none \\
ForgetNet & pred.\ hints & via hints \\
\Kone{} (here) & own state & none \\
stateless (here) & pred.\ hints & none \\
\bottomrule
\end{tabular}
\end{table}

What did we expect from each experiment, and what happened? Table~\ref{tab:registrations} gives, for each experiment that bears on this paper, what we expected to see (and how strongly), what would have shown us wrong, and what happened, failed predictions included.

\emph{The main comparisons.} Table~\ref{tab:stats} gives the main comparisons with two-sided 95\% Clopper--Pearson intervals (up to 15\% for $0/23$) and Holm's correction over them, a family we chose after the fact, since the predictions we recorded did not name primary tests. This adjustment does not account for the earlier choice of comparisons or seed-count extensions and is not a prospective family-wise error guarantee. The head-to-head includes its three pilot seeds. Excluding seeds 0--2 leaves 20 seeds, with $K{=}1$ ahead in 16, behind in 3 and tied in 1 on each band cell (two-sided sign $p=\PTPostPilotP{}$, exploratory). The reset and the clamp act on trained models and are reported per model, outside that family. Other tests are exploratory and unadjusted. The GIN comparison uses only the four seeds fitting both versions; the rates over all fitting seeds describe different, partly overlapping groups.

\emph{Checking our reimplementation.} Our reimplementation of the benchmark's rule had to approach the published ER accuracies before any head-to-head was read. A first run failed this check, with node accuracy at the model-free floor from a fault in our own code, and no claim was made from it. A second run passed at $n\le 160$ and fell short at $n{=}800$, a shortfall the degree input's tail explains (\S\ref{sec:trigger}).

\emph{A prediction that could not be tested.} The first version of the mechanism experiment described the failures as bounded reach and specified a reset triggered by a stall. No stall ever occurred, because every failing model reaches every node, so the reset could not have fired and the prediction could not have been tested. We record that it could not be tested. What the weights showed was a premature flood; the intervention was rewritten as a periodic reset with the period chosen after seeing the flood, and the confirmation with the period fixed in advance is the one in Table~\ref{tab:reset}.

\emph{The second algorithm.} We considered two further SALSA-CLRS tasks and set both aside before running either. Maximal independent set carries a size-invariant state, but its trajectory grows as $\log n$ and cannot exercise the depth axis. Minimum spanning tree (Prim) has a linear trajectory but a global argmin step that is not local, and Discrete NAR \citep{rodionov2025discrete}, which already removes the cross-step gradient, reports 100\% on it at every size. Neither would have tested the claim.

\emph{The first-page example.} Figure~\ref{fig:teaser}'s model and graph were chosen by a rule written down before the rollouts ran. Among seeds 3, 13 and 16, the seeds with both a flooding \bptt{} model and a saved \Kone{} model, we took the one whose median onset on blob16x64 is closest to the median onset of the six flooding models (ties to the lower seed), and the graph on which that model's onset is the median over the 20 graphs (the lower median, ties to the lower index). The rule gives seed 13 and the second graph of the set.

\emph{Grids and paths, and the linear probe.} The grids-and-paths runs trained three seeds per rule on short grids four nodes wide in place of the pool, and tested them on grid4x$l$ up to $l{=}512$ and on paths, chains of $n$ nodes, up to $n{=}1{,}024$. The linear probe regressed the step index $t$ on $h$ (ridge regression, scored on held-out graphs) along rollouts on training graphs and on blob16x64; an untrained GRU's state already encodes $t$ this way, so the probe cannot tell a counting state from a drifting one.

\begin{table*}[!htbp]
\caption{What we expected from each experiment, and what happened. Percentages are the probabilities we wrote down for each outcome before the run. ``Refuted by'' is the outcome that would defeat the claim the experiment was meant to support.}
\label{tab:registrations}
\centering\scriptsize\setlength{\tabcolsep}{4pt}
\begin{tabular}{p{0.16\textwidth}p{0.28\textwidth}p{0.20\textwidth}p{0.28\textwidth}}
\toprule
experiment & what we expected & refuted by & what happened \\
\midrule
Head to head (2 seeds, then 3, then 23) & \Kone{} matches \bptt{} within 0.02 everywhere (45\%); trails by 0.02--0.15 somewhere (35\%) & \Kone{} trails by more than 0.15 (20\%) & neither: matched on every cell but ER $n{=}800$, where \Kone{} was ahead. No predicted outcome covered this, so we retested at 23 seeds under new predictions, which held \\
Degree sweep & \bptt{} falls monotonically with degree while \Kone{} stays above 0.90 (50\%); both fall, \bptt{} faster (30\%) & neither falls & the second, but only in the band, not as a slope: both perfect at low degree, both on the floor at high degree \\
Truncation axis, $K\in\{1,2,4,8,\infty\}$ & a step at $K{=}1$ (40\%); monotone in $K$ (35\%) & flat in $K$ (25\%) & the step \\
Learning-rate control & \Kone{}'s advantage survives three learning rates at six seeds & \Kone{}'s advantage disappears at some rate & held: \Kone{} wins at every rate, the two rules' gradient norms are close, clipping is rare, and the runs at the operating rate reproduce the earlier ones \\
Deep families: grids and paths & the two rules separate on depth alone & they do not & they do not: both reach 508 hops on grids and 1,023 on paths in 3 of 3 seeds. The family was saturated \\
Deep families: blobs & the radius prediction at 3 seeds, then four at 23 seeds: \Kone{} perfect in at least 20 (60\%); \bptt{}'s rate below it (70\%); \bptt{}'s failures at a bounded radius; at least one \Kone{} failure (45\%) & fewer than 15 \Kone{} seeds perfect, or \bptt{}'s rate at or above \Kone{}'s & the radius prediction held; the first three held; the fourth did not (\Kone{} failed in no seed), and $0/23$ is reported as a bound \\
Reset with the period fixed in advance & \bptt{} to $\ge 22/23$ at $R{=}4$, every repaired radius 108 & the reset changes no radius & held: 22 of 23, every repaired radius 108; the exception, seed 3, floods by step 5 \\
$K{=}2$ on the deep family & floods at \bptt{}'s rate (55\%) & floods at \Kone{}'s rate & 7/23 failures against \bptt{}'s 6/23; paired $p=1$ \\
Linear probe for the step index in $h$ & descriptive only & none named & uninformative: an untrained GRU's state already encodes $t$ \\
Training pool & a deeper pool leaves \bptt{}'s flood rate where it is & \bptt{} floods in $\le 2/23$ under a deeper pool & refuted on all three pools \\
Depth versus density & a dense pool no deeper than the benchmark's own leaves the flood rate where it is (55\%); it removes the flood (30\%) & none named & it removes the flood, and 200 deep graphs also suffice (50\%) \\
Stateless block & as perfect as \Kone{} at the deep cells & it is not & held; its deficit on the band did not reach the level we had set ($p<0.05$ in the sign test; it is $p=0.13$) \\
Residual variant & \Kone{}'s immunity carries to another recurrence (45\%) & \Kone{} fails at a rate not distinguishable from \bptt{}'s & refuted, 23/23 against 23/23 \\
Degree clamp, after seeing the failures & density gates the flood and the clamp does nothing (50\%); the clamp repairs the models (30\%) & none named & the clamp repairs the models, 6 of 6 \\
Degree clamp, specified in advance & repairs at least 5 of 6 floods, band cells moving by less than 0.05 & fewer than 5 repaired & repairs 6 of 6, but the band prediction failed: the band cells rose by 0.2 to 0.6, to 1 for the GRU processor; partial for the residual variant \\
Processor trained without the feature & never floods and closes the band (45\%); still floods (35\%) & the band stays open (20\%) & never floods, band closed \\
Pool remedy on the residual variant & recurrence-independent (45\%) & it works only for the GRU processor & held, 23/23 and 23/23 under the shallow dense pool \\
Combined sample: the pool remedy on the benchmark's cells & no cell falls & any cell falls & none falls; the degree-25 and 33 cells rose from the floor to 1, an outcome we had given 25\% \\
\bottomrule
\end{tabular}
\end{table*}

\begin{table*}[!htbp]
\caption{What we expected from each experiment, and what happened (continued).}
\centering\scriptsize\setlength{\tabcolsep}{4pt}
\begin{tabular}{p{0.16\textwidth}p{0.28\textwidth}p{0.20\textwidth}p{0.28\textwidth}}
\toprule
experiment & what we expected & refuted by & what happened \\
\midrule
Counting or drift, on the saved weights of 12 models & mixed (35\%), drift (35\%), counting (30\%); the control holds in at least 5 of the 6 failing models (80\%) & a drift outcome, for counting as the mechanism & mixed: no model counts, three drift and three lie between; the control held in 6 of 6 and every model reproduced its recorded results, so the paper does not describe the flood as counting \\
Onset against a uniform degree input & each failing model first floods above the pool's 99th percentile (65\%); its onset does not rise with the input (80\%) & a failing model flooding at 1.3 or below & both: first floods at 1.4 to 2.0, onsets falling with the input in 6 of 6 \\
Healthy models under the same input & a delayed flood up to 3.0 in at least one \Kone{} model (45\%), in none (30\%), or only immediate floods or stalls (25\%); in at least one healthy \bptt{} model (65\%) & a delayed flood in a \Kone{} model, for \Kone{}'s protection & all three \Kone{} models and two of three healthy \bptt{} models: \Kone{} changes how far out the input must be \\
Mazes at the blob input & at least 4 of the 6 failing models flood on a maze with \xdeg{} at 1.9 (60\%); at most 1 (25\%) & at most 1, for the input's value as the trigger & 6 of 6 \\
Healthy models under the switch and transplant & none floods (85\%) & none named & none floods \\
Progressive loss under \bptt{}, 23 seeds & at least 4 seeds fail, as floods (55\%); 1 to 3 (30\%); none (15\%) & either remedy at 0/23 & the first: 5 of 23 fail, all as floods; all runs have low logged training loss \\
Random initial states under \bptt{}, 23 seeds & at least 4 fail (40\%); 1 to 3 (30\%); none (30\%) & either remedy at 0/23 & 1 of 23, a flood; all runs have low logged training loss \\
Sum aggregation without the degree input, 23 seeds per rule & \bptt{} fails in at least 5 seeds (45\%), \Kone{} in at least 5 (25\%); at least two thirds of the failures flood (65\%) & 5 or more flooding seeds under \bptt{}, for the feature as the only trigger & 15 and 10 of 23 fail, against the means' 0, and 7 flood under \bptt{}, so the feature is not the only trigger; but 10 flood only, 14 stall only and one does both; both rules fail ($p=0.30$) \\
\bottomrule
\end{tabular}
\end{table*}

\begin{table*}[!htbp]
\caption{What we expected from each experiment, and what happened (continued).}
\centering\scriptsize\setlength{\tabcolsep}{4pt}
\begin{tabular}{p{0.16\textwidth}p{0.28\textwidth}p{0.20\textwidth}p{0.28\textwidth}}
\toprule
experiment & what we expected & refuted by & what happened \\
\midrule
The sums brought back into range, on the saved weights & at the pool's median count at least 20 of the 25 failing models are exact (65\%); failures begin only above the pool's 95th percentile (60\%) & more than 5 models failing at a count of 7, for the range as the cause & 17 of 25 repaired, but 12 fail at 7: the range is not the whole account \\
The same, further predictions & at five, at least 19 of 21 healthy models stay exact (85\%); further out, floods come no later and stalls grow no smaller (70\% and 60\%); at thirty, at least 11 healthy models fail (55\%) and at least 20 of the 25 keep their type (70\%); capping the sums at eleven helps the dense ER cells in at least 18 of 25 (60\%) & none named & 19 of 21 stay exact; the predicted flood trend held in 10 of 11 models, but the predicted stall trend held in only 6 of 14; at thirty, 13 healthy models fail on blob16x64 and 14 on either deep cell (Table~\ref{tab:sum281}); only 15 of 25 keep their type; the cap helps in 8 of 25, all \Kone{}, and hurts most \bptt{} models \\
The readout of untouched nodes, degree input, 6 failing models & with the true reach mask fed in, its median climbs over steps 2 to 20 at $v{=}1.9$, and at most a quarter as fast at 1.0, in at least 5 of 6 models (70\%) & none named; descriptive & held in 6 of 6, at most a fifth as fast \\
The same for the 25 failing sum models & floods: the untouched nodes' readout climbs at 30 neighbors' worth and at most a quarter as fast at five, in at least 8 of 11 (60\%); stalls: the readout of nodes due at the step falls at 30 and at most a quarter as fast at five, in at least 10 of 14 (50\%) & none named; descriptive & held for the floods, 10 of 11; failed for the stalls, 2 of 14, so the stalls stay unexplained \\
The sum models with the training graphs' self-loops & the failing models still fail (65\%); at five neighbors' worth at least 20 of 25 are exact (50\%); at seven at most 5 fail (45\%); at least 19 of 21 healthy models stay exact (70\%) & 13 or more exact under the self-loops, for the missing self-loops as the cause & none of the 25 repaired, so the failures are not an artifact of the missing self-loops; but only 15 are exact at five (17 without them), the same 12 fail at seven, and 4 of the 21 healthy models break \\
The benchmark's own models (GIN, PGN and RecGNN as released; GIN with sums), 10 seeds each & the three released models reproduce their published accuracy (65\%); in each, at most 1 fitting seed floods at blob16x64 (40\%), or 3 or more in one of them (35\%); GIN with sums floods in at least 3 (55\%) and more often than GIN in at least 7 paired seeds (55\%); at most 1 per model floods on maze32 (55\%) & a released model flooding in 3 or more seeds, for our account of the trigger in models of that kind & all three reproduce; GIN floods in 3 of 7 fitting seeds, PGN in 1 of 10, RecGNN in 0 of 8 (one seed timed out), each flooding seed on the mazes too, from nearly the same step; GIN with sums declares nodes early in 7 of 7; only 4 seeds fit in both GIN models, too few for the paired predictions. Our account of the trigger covers our processor only \\
Which sum carries the failure, on the saved weights of the 46 sum models & at five neighbors' worth each sum alone repairs fewer of the 25 than both together (70\%), and the hidden-state sum more than the reach-mask sum (55\%); of the 12 failing at seven with both, at least 8 also fail with each sum alone (60\%) & none named; descriptive & held: the hidden-state sum alone repairs 15, the reach-mask sum alone none, both 17; failed: 7 of the 12 \\
The drift probe for the 11 flooding sum models, the count as the dial & no flooding model counts (65\%); the positive control holds in at least 80\% of them (75\%); no healthy control floods (80\%) & a model that counts & refuted: 4 count, 3 drift, 4 lie between; the control held in 11 of 11; 1 of 6 controls floods \\
The benchmark's own models on its own test sets not read step by step before (Watts--Strogatz, and ER and Delaunay at 1,600 nodes) & GIN seed 42 floods on Delaunay 1,600 (65\%) and no other unmodified seed does (60\%); none floods on the other sets (70 and 85\%); GIN with sums floods on every set above 16 nodes (75\%) & none named; what the paper would say was fixed beforehand for either outcome & held: GIN seed 42 floods in 199 of 200 graphs, and no other unmodified seed declares a node early on any set; failed by one seed: GIN with sums in 6 of 7 on Watts--Strogatz 80 \\
\bottomrule
\end{tabular}
\end{table*}

\begin{table*}[!htbp]
\caption{The paper's main comparisons. Counts are seeds out of those trained: failing seeds (floods or stalls) in the rows on failures, and flooding seeds in the rows on floods, where every failing seed floods; for the benchmark's GIN, seeds that declare nodes early among the four seeds fitting both versions. Intervals are two-sided 95\% Clopper--Pearson intervals, and in parentheses are the seeds that fail under only the first or only the second choice. The $p$ values are two-sided: exact McNemar tests for paired counts, including the four shared fitting GIN seeds, and exact sign tests for node accuracy. Holm $p$: Holm's correction over the eight comparisons, a family chosen after the fact. The interventions act on trained models, each its own control, and are reported as models repaired and broken, outside the family.}
\label{tab:stats}
\centering\scriptsize\setlength{\tabcolsep}{3pt}
\begin{tabular}{p{0.30\textwidth}lrlrr}
\toprule
comparison & where & counts [95\% interval] (discordant) & test & $p$ & Holm $p$ \\
\midrule
Floods, \bptt{} against \Kone{} (GRU processor) & \S\ref{sec:flood} & 6/23 [0.10, 0.48] vs.\ 0/23 [0.00, 0.15] (6, 0) & McNemar & 0.031 & 0.156 \\
Failures, sums against means, \bptt{} & \S\ref{sec:sums} & 15/23 [0.43, 0.84] vs.\ 0/23 [0.00, 0.15] (15, 0) & McNemar & $<$0.001 & $<$0.001 \\
Failures, sums against means, \Kone{} & \S\ref{sec:sums} & 10/23 [0.23, 0.66] vs.\ 0/23 [0.00, 0.15] (10, 0) & McNemar & 0.002 & 0.012 \\
Failures with sums, \bptt{} against \Kone{} & \S\ref{sec:sums} & 15/23 [0.43, 0.84] vs.\ 10/23 [0.23, 0.66] (10, 5) & McNemar & 0.302 & 0.604 \\
Floods, progressive loss against \bptt{} & \S\ref{sec:remedies} & 5/23 [0.07, 0.44] vs.\ 6/23 [0.10, 0.48] (1, 2) & McNemar & 1 & 1 \\
Floods, random initial states against \bptt{} & \S\ref{sec:remedies} & 1/23 [0.00, 0.22] vs.\ 6/23 [0.10, 0.48] (0, 5) & McNemar & 0.062 & 0.250 \\
ER 800 node accuracy, \Kone{} against \bptt{} (seeds up, down) & \S\ref{sec:headtohead} & 19 vs.\ 3 & sign & $<$0.001 & 0.006 \\
Benchmark's GIN, sums against max (shared fitting seeds) & \S\ref{sec:benchmark} & 4/4 [0.40, 1.00] vs.\ 0/4 [0.00, 0.60] (4, 0) & McNemar & 0.125 & 0.375 \\
\midrule
\multicolumn{6}{l}{\emph{Interventions on trained models, outside the family}} \\
Reset every four steps, \bptt{} models & \S\ref{sec:mechanism} & 5 of 6 failing repaired, 0 of 17 broken & -- & -- & -- \\
Degree clamp at 1.0, \bptt{} models & \S\ref{sec:trigger} & 6 of 6 failing repaired, 0 of 17 broken & -- & -- & -- \\
\bottomrule
\end{tabular}
\end{table*}

\begin{table*}[!htbp]
\caption{The benchmark's own models, trained with its code, configurations and hints, ten seeds each (RecGNN nine: one seed hit the time limit), and read on our families. Reproduced: mean $\pi$ node accuracy of at least 0.98 on the benchmark's three 16-node test sets and within five points of the mean over its fifteen published for models trained with hints \citep{minder2023salsaclrs}. Fitting: early and late-or-never fractions each below 1\% on our ER~16 set; the classes count fitting seeds only. F/S/M/E: flood, stall, mixed, exact; the exact class has both timing-error fractions below 1\%. Early: fitting seeds with at least 1\% of nodes declared early, with a two-sided 95\% Clopper--Pearson interval conditional on fitting. Last column: mean $\pi$ node accuracy of the fitting seeds, with the lowest-index floor in brackets; these models predict $\pi$ with their own output head, so it is not decoded from timing as ours is. Their $\pi$ is poor on the deep graphs even for seeds in the exact reach class, for a reason we do not study.}
\label{tab:salsa}
\centering\footnotesize\setlength{\tabcolsep}{1.4pt}
\begin{tabular}{lrrrrrrrr}
\toprule
& & \multicolumn{2}{c}{node accuracy} & & \multicolumn{2}{c}{blob16x64} & maze32 & $\pi$ node acc. \\
\cmidrule(lr){3-4}\cmidrule(lr){6-7}
model & reproduced & 16-node sets & 15 sets & fitting & F/S/M/E & early & F/S/M/E & blob16x64 (floor) \\
\midrule
GIN (max), as released & yes & 0.982 & 0.787 & 7/10 & 3/1/0/3 & 3 [0.10, 0.82] & 3/2/0/2 & 0.21 (0.19) \\
PGN (max), as released & yes & 1.000 & 0.889 & 10/10 & 1/0/0/9 & 1 [0.00, 0.45] & 1/0/0/9 & 0.34 (0.19) \\
RecGNN (max), as released & yes & 1.000 & 0.853 & 8/9 & 0/0/0/8 & 0 [0.00, 0.37] & 0/0/0/8 & 0.48 (0.19) \\
GIN with sums & -- & 0.993 & 0.651 & 7/10 & 6/0/1/0 & 7 [0.59, 1.00] & 3/0/3/1 & 0.27 (0.19) \\
\bottomrule
\end{tabular}
\end{table*}

\begin{table*}[!htbp]
\caption{When the benchmark's unmodified flooding seeds flood: the median step of the first early declaration over the graphs that have one, its range in brackets, and in parentheses the graphs with an early declaration out of those read. A seed floods only on graphs whose search runs past its onset. Zero counts on shorter trajectories do not test behavior beyond that horizon.}
\label{tab:salsa_onsets}
\centering\footnotesize\setlength{\tabcolsep}{3pt}
\begin{tabular}{lrrrrr}
\toprule
flooding seed & blob16x32 & blob16x64 & maze16 & maze32 & Delaunay 1,600 \\
\midrule
GIN seed 42 & 18 [18--19] (20/20) & 18 [18] (20/20) & 18 [18] (40/40) & 18 [18] (20/20) & 18 [18--19] (199/200) \\
GIN seed 43 & 40 [40--43] (7/20) & 39 [39--43] (20/20) & 39 [38--43] (40/40) & 39 [38--40] (20/20) & -- \\
GIN seed 46 & -- (0/20) & 94 [94--95] (4/20) & 98 [95--135] (15/40) & 94 [94--102] (20/20) & -- \\
PGN seed 46 & 28 [27--28] (13/20) & 28 [28] (20/20) & 30 [29--36] (29/40) & 28 [27--38] (20/20) & -- \\
\bottomrule
\end{tabular}
\end{table*}

\begin{table*}[!htbp]
\caption{The benchmark's own models on its own test sets that we had not read step by step (first 200 graphs each; WS: Watts--Strogatz). Each cell: fitting seeds that declare at least 1\% of the nodes reached early, of the fitting seeds. The first row gives each set's mean trajectory length; every flood of the unmodified models has its first early declaration at step 18 or later (Table~\ref{tab:salsa_onsets}). Zero counts on shorter trajectories do not establish behavior beyond their observed horizons.}
\label{tab:p283}
\centering\footnotesize\setlength{\tabcolsep}{4pt}
\begin{tabular}{lrrrrrrr}
\toprule
 & WS 16 & WS 80 & WS 160 & WS 800 & WS 1,600 & ER 1,600 & Delaunay 1,600 \\
\midrule
mean trajectory length & 3.7 & 7.2 & 8.0 & 12.9 & 9.9 & 5.4 & 23.3 \\
\midrule
GIN & 0/7 & 0/7 & 0/7 & 0/7 & 0/7 & 0/7 & 1/7 \\
PGN & 0/10 & 0/10 & 0/10 & 0/10 & 0/10 & 0/10 & 0/10 \\
RecGNN & 0/8 & 0/8 & 0/8 & 0/8 & 0/8 & 0/8 & 0/8 \\
GIN with sums & 1/7 & 6/7 & 7/7 & 7/7 & 7/7 & 7/7 & 7/7 \\
\bottomrule
\end{tabular}
\end{table*}

\begin{figure}[!htbp]
\centering\includegraphics[width=0.62\textwidth]{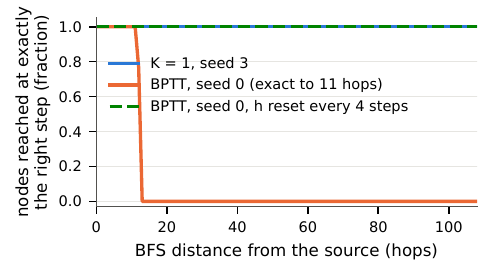}
\caption{Fraction of nodes at each BFS distance that are declared reached at exactly the right step, on blob16x64. A \Kone{} model is exact to the last node; a flooding \bptt{} model is exact to eleven hops and then declares everything reached; the same weights with $h$ zeroed every four steps are exact to the last node.}
\label{fig:flood}
\end{figure}

\begin{table*}[!htbp]
\caption{The degree input clamped at inference, on all 23 seeds of both rules, for the GRU processor and its residual variant, over the 21 cells of the combined sample. This is an intervention on trained models: the same weights and the same graphs as the standard (std.) columns, with the degree input overwritten at inference. ``Cells exact'' counts the cells on which every seed is at graph accuracy 1. ``Radius lost'' is the largest loss of exact-timing radius on any blob or maze set. For the GRU processor either clamp makes every model exact on every cell; for the residual variant it closes the ER cells but repairs the deep family only partly and breaks the mazes.}
\label{tab:clamp}
\centering\footnotesize\setlength{\tabcolsep}{3pt}
\begin{tabular}{lrrrrrrrrrrr}
\toprule
& \multicolumn{3}{c}{perfect at blob 32 \& 64} & \multicolumn{3}{c}{cells exact (of 21)} & radius lost & \multicolumn{2}{c}{ER 800 node} & \multicolumn{2}{c}{Delaunay 800 node} \\
\cmidrule(lr){2-4}\cmidrule(lr){5-7}\cmidrule(lr){9-10}\cmidrule(lr){11-12}
model, rule (23 seeds) & std.\ & $x_{\mathrm{deg}}{=}1.0$ & $0.5$ & std.\ & $1.0$ & $0.5$ & (max, deep sets) & std.\ & $1.0$ & std.\ & $1.0$ \\
\midrule
GRU, BPTT & 17/23 & 23/23 & 23/23 & 9 & 21 & 21 & 0 & 0.581 & 1.000 & 1.000 & 1.000 \\
GRU, $K{=}1$ & 23/23 & 23/23 & 23/23 & 13 & 21 & 21 & 0 & 0.771 & 1.000 & 1.000 & 1.000 \\
residual, BPTT & 0/23 & 0/23 & 10/23 & 2 & 6 & 4 & 534 & 0.379 & 0.998 & 0.439 & 0.381 \\
residual, $K{=}1$ & 0/23 & 11/23 & 16/23 & 2 & 9 & 11 & 533 & 0.653 & 1.000 & 0.781 & 0.749 \\
\bottomrule
\end{tabular}
\end{table*}

\paragraph{Reading the sum models' failures.} Of the sums of \S\ref{sec:sums} rescaled at inference, the reach-mask sum alone breaks all 21 healthy models, so it cannot show whether that sum contributes (Table~\ref{tab:sum281}). With both sums rescaled, 8 of the 25 failing models still fail at the training median and 12 at $c{=}7$, below the pool's 95th percentile of 9 (Table~\ref{tab:range}). Adding the training graphs' self-loops to the blobs repairs none of the 25. Under large sums the readout of untouched nodes climbs in 10 of the 11 flooding models, as under a tail degree input, but the switch and transplant of \S\ref{sec:drift}, driven by the neighbor count instead of the degree input, give no single answer: on the switch, 4 of the 11 count, 3 drift and 4 lie between.

\begin{table}[!htbp]
\caption{The 25 sum models that fail at the deep cells, run from their saved weights with both neighborhood sums rescaled at inference to $c$ neighbors' worth ($c$ times the neighborhood mean). Columns: the failure type under the true count; exact at $c{=}5$; $c_f$, the smallest $c\ge 5$ of the grid at which the model fails at blob16x64; failing at $c{=}7$; the type at $c{=}30$. The training pool's count has median 5, 95th percentile 9 and maximum 15; a blob node has 15 to 18 neighbors.}
\label{tab:range}
\centering\footnotesize\setlength{\tabcolsep}{3pt}
\begin{tabular}{lrlrrrl}
\toprule
& & unscaled & \multicolumn{4}{c}{rescaled count $c$} \\
\cmidrule(lr){4-7}
rule & seed & type & exact at 5 & first failure $c_f$ & fails at 7 & type at 30 \\
\midrule
\bptt{} & 0 & stall & no & 5 & yes & exact \\
\bptt{} & 1 & flood & yes & 13 & no & flood \\
\bptt{} & 2 & flood & yes & 15 & no & flood \\
\bptt{} & 3 & flood & yes & 11 & no & flood \\
\bptt{} & 6 & stall & no & 5 & yes & flood \\
\bptt{} & 8 & mixed & no & 5 & yes & flood \\
\bptt{} & 9 & stall & yes & 11 & no & flood \\
\bptt{} & 12 & stall & no & 5 & yes & flood \\
\bptt{} & 13 & flood & yes & 7 & yes & flood \\
\bptt{} & 14 & flood & yes & 9 & no & flood \\
\bptt{} & 15 & flood & no & 5 & yes & flood \\
\bptt{} & 16 & stall & no & 5 & yes & flood \\
\bptt{} & 19 & stall & no & 5 & yes & flood \\
\bptt{} & 20 & stall & yes & 13 & no & stall \\
\bptt{} & 21 & flood & yes & 7 & yes & flood \\
\midrule
\Kone{} & 1 & flood & yes & 15 & no & flood \\
\Kone{} & 4 & stall & yes & 9 & no & stall \\
\Kone{} & 6 & stall & yes & 13 & no & stall \\
\Kone{} & 8 & stall & yes & 11 & no & exact \\
\Kone{} & 10 & stall & yes & 7 & yes & flood \\
\Kone{} & 11 & stall & no & 5 & yes & flood \\
\Kone{} & 12 & flood & yes & 7 & yes & flood \\
\Kone{} & 13 & flood & yes & 9 & no & flood \\
\Kone{} & 17 & stall & yes & 15 & no & stall \\
\Kone{} & 18 & stall & yes & 11 & no & stall \\
\bottomrule
\end{tabular}
\end{table}

\begin{table}[!htbp]
\caption{The 46 sum models (25 failing and 21 healthy at the deep cells), run from their saved weights with one neighborhood sum or both rescaled at inference to $c$ neighbors' worth; a sum not rescaled keeps its true count. Each cell: failing models made exact / healthy models broken.}
\label{tab:sum281}
\centering\footnotesize\setlength{\tabcolsep}{4pt}
\begin{tabular}{lrrr}
\toprule
rescaled at inference & $c{=}5$ & $c{=}7$ & $c{=}30$ \\
\midrule
hidden-state sum only & 15 / 1 & 16 / 1 & 0 / 21 \\
reach-mask sum only & 0 / 21 & 0 / 20 & 7 / 8 \\
both sums & 17 / 2 & 13 / 2 & 2 / 14 \\
\bottomrule
\end{tabular}
\end{table}

\begin{table}[!htbp]
\caption{The reset, specified before it was run: $h\leftarrow 0$ every $R$ steps at inference, $f$ kept, the same $R$ for every seed. This is an intervention on already-trained models. The 46 models of Table~\ref{tab:rates}'s \bptt{} and \Kone{} rows are measured again on the same graphs, with one thing changed at inference.}
\label{tab:reset}
\centering\footnotesize\setlength{\tabcolsep}{3pt}
\begin{tabular}{lrrrr}
\toprule
& \multicolumn{3}{c}{perfect at blob 32 \& 64} & radius lost \\
\cmidrule(lr){2-4}
rule (23 seeds) & none & $R{=}4$ & $R{=}8$ & (max, any set) \\
\midrule
full BPTT & 17/23 & 22/23 & 22/23 & 0 \\
$K=1$ & 23/23 & 23/23 & 23/23 & 0 \\
\bottomrule
\end{tabular}
\end{table}

\paragraph{Reading the residual variant's failures.} The residual variant of \S\ref{sec:remedies} fails at both deep cells in 23 of 23 seeds under both rules. Three failing seeds per rule were re-run from the saved weights on eight blob16x64 graphs (Table~\ref{tab:probe232}). Every failure is a flood: 88 to 98\% of nodes are declared reached early and none late. Zeroing $h$ every two steps makes all three \Kone{} models exact to the last node, so their floods are again carried in the state. The \bptt{} floods begin at step two or three and survive resets every two or four steps, so this variant's \bptt{} failures follow neither route. On the deep cells \Kone{}'s exact-timing radius exceeds \bptt{}'s in 23 of 23 seeds. The same variant fails the mazes in 1 of 23 seeds under \bptt{} and 6 under \Kone{}. Without the degree feature it still fails, in 8 of 23 seeds under \bptt{} and 4 under \Kone{}, and clamping the feature repairs it only in part (Table~\ref{tab:clamp}).

\begin{table}[!htbp]
\caption{The residual-variant probe: three failing seeds per rule re-run from the saved weights on eight blob16x64 graphs. ``Early'' and ``late'' are fractions of nodes declared reached before and after the correct step; every model reaches every node.}
\label{tab:probe232}
\centering\footnotesize\setlength{\tabcolsep}{3pt}
\begin{tabular}{lrrrrr}
\toprule
& \multicolumn{3}{c}{standard rollout} & \multicolumn{2}{c}{radius, reset} \\
\cmidrule(lr){2-4}\cmidrule(lr){5-6}
residual model & radius & early & late & $R{=}2$ & $R{=}4$ \\
\midrule
BPTT, seed 0 & 1 & 0.976 & 0.000 & 1 & 1 \\
BPTT, seed 1 & 2 & 0.975 & 0.000 & 2 & 2 \\
BPTT, seed 9 & 2 & 0.975 & 0.000 & 2 & 2 \\
$K{=}1$, seed 0 & 6 & 0.901 & 0.000 & 108 & 108 \\
$K{=}1$, seed 1 & 3 & 0.952 & 0.000 & 108 & 3 \\
$K{=}1$, seed 9 & 7 & 0.885 & 0.000 & 108 & 108 \\
\bottomrule
\end{tabular}
\end{table}

\paragraph{The stateless block.} The stateless block of \S\ref{sec:remedies} and ForgetNet \citep{bohde2024markov} both drop the carried state, and ours also passes no gradient through the fed-back reach mask (Table~\ref{tab:positioning}). ForgetNet was evaluated on CLRS's 30 algorithms at $n{=}64$, where a late flood has little room to show. On the band, where \Kone{} leads \bptt{} by 0.19 and 0.31 in the samples of Table~\ref{tab:rates} (0.15 and 0.22 on the sample of Table~\ref{tab:headtohead}), the stateless block lies between the two rules: above \bptt{} by 0.11 and 0.17 node accuracy (exact sign test over seeds, $p<0.001$ in each cell) and below \Kone{} by 0.08 and 0.14, a gap 23 seeds do not resolve ($p=0.13$ in each).

\begin{table}[!htbp]
\caption{The trigger, on the saved weights of \S\ref{sec:flood}: exact-timing radius on newly generated chains of 64 cliques of $q$ nodes (mean degree 4.0, 7.5, 15.2 and eccentricity 68, 83, 91 for $q{=}4,8,16$; largest distances 87, 102, 106), the fraction of nodes reached early at $q{=}16$, and blob16x64 with the degree input set to 1.0 at inference (0.5 gives the same). These chains are a separate sample of eight graphs per size, each with its own random source, so their depths differ from blob16x64's. Every model reproduces every training hint on every step of all 6,000 training graphs.}
\label{tab:density}
\centering\footnotesize\setlength{\tabcolsep}{2.5pt}
\begin{tabular}{lrrrrrr}
\toprule
& \multicolumn{3}{c}{$q$-clique chains} & early & \multicolumn{2}{c}{$x_{\mathrm{deg}}{=}1.0$} \\
\cmidrule(lr){2-4}\cmidrule(lr){6-7}
model & $q{=}4$ & $8$ & $16$ & at 16 & radius & early \\
\midrule
BPTT, seed 3 & 87 & 102 & 5 & 0.92 & 108 & 0.00 \\
BPTT, seed 0 & 87 & 102 & 11 & 0.81 & 108 & 0.00 \\
BPTT, seed 13 & 87 & 102 & 16 & 0.75 & 108 & 0.00 \\
BPTT, seed 16 & 87 & 102 & 21 & 0.58 & 108 & 0.00 \\
BPTT, seed 22 & 87 & 102 & 23 & 0.59 & 108 & 0.00 \\
BPTT, seed 2 & 87 & 102 & 25 & 0.36 & 108 & 0.00 \\
healthy (6 models) & 87 & 102 & 106 & 0.00 & 108 & 0.00 \\
\bottomrule
\end{tabular}
\end{table}

\begin{table}[!htbp]
\caption{Exact-timing radius of the six \bptt{} models that fail at blob16x64, on the four blob sizes, and the share of blob16x64's nodes each declares reached early. The radius is pooled over the graphs of a set, so the graphs that flood earliest set it; a radius that stops growing with the graph marks a flood that begins at a model-specific step. Every \Kone{} model reaches the set's largest distance at every size.}
\label{tab:failing}
\centering\footnotesize\setlength{\tabcolsep}{2.5pt}
\begin{tabular}{lrrrrr}
\toprule
& \multicolumn{4}{c}{radius at $l$ cliques (mean ecc.)} & early \\
\cmidrule(lr){2-5}
\bptt{} model & 8 (11) & 16 (21) & 32 (41) & 64 (77) & at 64 \\
\midrule
seed 3 & 5 & 5 & 5 & 5 & 0.92 \\
seed 0 & 12 & 11 & 11 & 11 & 0.81 \\
seed 13 & 15 & 16 & 16 & 16 & 0.74 \\
seed 16 & 15 & 28 & 26 & 19 & 0.55 \\
seed 22 & 15 & 28 & 26 & 24 & 0.57 \\
seed 2 & 15 & 28 & 33 & 25 & 0.22 \\
\midrule
$K{=}1$, any seed & 15 & 28 & 57 & 108 & 0.00 \\
\bottomrule
\end{tabular}
\end{table}

\begin{figure}[!htbp]
\centering\includegraphics[width=0.62\textwidth]{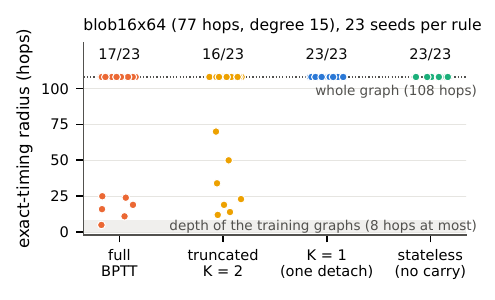}
\caption{Exact-timing radius per seed on blob16x64 (77 hops deep on average, degree 15) for four training rules under the pool. A radius of 108, the set's largest distance, meets the 95\% timing criterion at every distance. \Kone{} and the stateless block never fall below it; \bptt{} and $K{=}2$ do in about a quarter of seeds.}
\label{fig:rates}
\end{figure}

\paragraph{The drift probe.} The twelve models of \S\ref{sec:drift} are the saved weights of the six failing \bptt{} seeds, three healthy \bptt{} seeds and three \Kone{} seeds. Every one reproduces its recorded standard results on blob16x32 and blob16x64, and a transplant at $k{=}0$ reproduces the standard rollout node for node.
\begin{itemize}\setlength{\itemsep}{1pt}\setlength{\parsep}{0pt}
\item \emph{Onset.} A graph's onset $o$ is the first step $t$ at which a node at true distance greater than $t$ is declared reached, read as the graph's largest distance if there is none; $t^*$ is the graph's onset under the true input.
\item \emph{Switch.} The degree input is 1.0 for steps 1 to $k$ and the true input after, with $h$ and $f$ carried, for $k\in\{2,4,8,16,24,32,48,64\}$.
\item \emph{Transplant.} A fresh rollout under the true input starts with only the source reached, and every other node starts from a state of the rollout with the input at 1.0, taken at step $k$: its own, if it is still unreached there and at least $k{+}3$ hops from the source, and otherwise that of a random such node of the same graph and degree.
\item \emph{Statistic.} For a graph and a $k$,
\begin{align}
s_{\mathrm{switch}} &= \frac{(k+t^*)-o}{\min(k,\,t^*-1)}, \label{eq:s_switch}\\
s_{\mathrm{transplant}} &= \frac{t^*-o}{\min(k,\,t^*-1)}, \label{eq:s_transplant}
\end{align}
the flood's observed advance on drift's prediction as a share of the advance counting predicts: 1 under counting and 0 under drift. It is used where $t^*\ge 2$ and, for the switch, where $k{+}t^*$ is below the graph's largest distance, so that both predictions can be observed. It is negative when the flood comes later than drift predicts and above 1 when it comes earlier than counting predicts.
\item \emph{Control.} $c_P=(t^*-o)/k$, for the transplant from the rollout under the true input with $k\le t^*{-}2$, is 1 under both readings.
\item \emph{Classes.} A model counts if its median of $s$ over graphs and $k$ is at least 0.75, drifts if it is at most 0.25, and is mixed otherwise. The result is counting if at least four of the six failing models count on the switch and most of those that pass the control count on the transplant, drift if the same holds with drift in place of counting, and mixed otherwise.
\end{itemize}
Before any model was read, synthetic onsets built from each prediction recovered their class. Defining the onset as the first step with 1\% of the nodes early gives the same classes; with 0.5 as the typical value the switch statistic is lower in five models and equal in the sixth. In the sweep a model floods at $v$ when at least 1\% of the nodes are declared early, and on maze32 the rollouts stop at 150 steps. In Figure~\ref{fig:drift}b a delay above the dashed line is a flood later than drift predicts, and a delay below zero a flood earlier than under the true input throughout.

\begin{table*}[!htbp]
\caption{The drift probe per model, on 20 blob16x64 graphs. $t^*$: median onset of the flood under the true input. Switch and transplant: the model's median, over graphs and $k$, of the statistic $s$ of Eqs.~(\ref{eq:s_switch}) and (\ref{eq:s_transplant}) (Appendix~\ref{app:registrations}), the flood's observed advance on drift's prediction as a share of the advance counting predicts: 0 under drift, 1 under counting. $c_P$: the control, the flood's advance as a fraction of $k$ when the transplanted state comes from the rollout under the true input, 1 under both readings. Reading: drifts at 0.25 or below, counts at 0.75 or above, mixed between. First flood: the smallest uniform degree input $v$ at which at least 1\% of the nodes are declared early. Maze32 early: the fraction of maze32's nodes declared early within 150 steps with the degree input at 1.9.}
\label{tab:drift}
\centering\footnotesize\setlength{\tabcolsep}{4pt}
\begin{tabular}{lrrrrlrr}
\toprule
& onset & \multicolumn{2}{c}{share of typical steps that count} & & reading & first flood & maze32 early \\
\cmidrule(lr){3-4}
model & $t^*$ & switch & transplant & control & & at $v$ & at $v{=}1.9$ \\
\midrule
BPTT seed 3 & 4 & 0.33 & 0.33 & 1.00 & mixed & 1.4 & 0.99 \\
BPTT seed 0 & 10 & 0.40 & 0.44 & 1.00 & mixed & 1.9 & 0.97 \\
BPTT seed 13 & 16 & 0.27 & 0.29 & 1.00 & mixed & 1.6 & 0.96 \\
BPTT seed 16 & 24.5 & 0.12 & 0.17 & 1.00 & drifts & 2.0 & 0.91 \\
BPTT seed 22 & 23 & 0.11 & 0.18 & 1.00 & drifts & 1.9 & 0.92 \\
BPTT seed 2 & 42 & $-$0.38 & 0.25 & 1.00 & drifts & 2.0 & 0.90 \\
\midrule
BPTT seed 4 & -- & -- & -- & -- & healthy & 2.8 & 0.00 \\
BPTT seed 5 & -- & -- & -- & -- & healthy & none & 0.00 \\
BPTT seed 6 & -- & -- & -- & -- & healthy & 2.6 & 0.00 \\
$K{=}1$ seed 3 & -- & -- & -- & -- & healthy & 2.8 & 0.00 \\
$K{=}1$ seed 13 & -- & -- & -- & -- & healthy & 2.9 & 0.00 \\
$K{=}1$ seed 16 & -- & -- & -- & -- & healthy & 2.2 & 0.34 \\
\bottomrule
\end{tabular}
\end{table*}

\begin{table*}[!htbp]
\caption{Median flood onset on blob16x64 after the first $k$ steps under the typical input (the switch), for the six failing \bptt{} models, against the two predictions: drift, $k+t^*$; counting, $\max(t^*,k+1)$.}
\label{tab:drift_onsets}
\centering\footnotesize\setlength{\tabcolsep}{4pt}
\begin{tabular}{lrrrrrrrrrr}
\toprule
& & \multicolumn{3}{c}{$k{=}2$} & \multicolumn{3}{c}{$k{=}8$} & \multicolumn{3}{c}{$k{=}32$} \\
\cmidrule(lr){3-5}\cmidrule(lr){6-8}\cmidrule(lr){9-11}
model & $t^*$ & observed & drift & count & observed & drift & count & observed & drift & count \\
\midrule
BPTT seed 3 & 4 & 6 & 6 & 4 & 11 & 12 & 9 & 35 & 36 & 33 \\
BPTT seed 0 & 10 & 12 & 12 & 10 & 17 & 18 & 10 & 38 & 42 & 33 \\
BPTT seed 13 & 16 & 18 & 18 & 16 & 22 & 24 & 16 & 40 & 48 & 33 \\
BPTT seed 16 & 24.5 & 30 & 26.5 & 24.5 & 31 & 32.5 & 24.5 & 53 & 56.5 & 33 \\
BPTT seed 22 & 23 & 33 & 25 & 23 & 33 & 31 & 23 & 50 & 55 & 33 \\
BPTT seed 2 & 42 & 50 & 44 & 42 & 46 & 50 & 42 & 70.5 & 74 & 42 \\
\bottomrule
\end{tabular}
\end{table*}

\begin{table*}[!htbp]
\caption{The literature's remedies added to \bptt{}, 23 seeds each, on the combined sample. $\alpha$: the weight of the progressive loss's extra rollout. Failing seeds: seeds not at graph accuracy 1 at both deep blob cells; every one floods. Exact McNemar $p$ against the 6 of 23 of \bptt{} and the 0 of 23 of \Kone{}, on the same seeds. Paired with \bptt{} by seed: seeds repaired and broken. Node accuracy: the mean paired difference from \bptt{}, with the numbers of seeds above and below.}
\label{tab:remedies}
\centering\footnotesize\setlength{\tabcolsep}{4pt}
\begin{tabular}{lrrrrrrr}
\toprule
& failing & \multicolumn{2}{c}{McNemar $p$ against} & \multicolumn{2}{c}{paired with \bptt{}} & \multicolumn{2}{c}{node acc.\ against \bptt{}} \\
\cmidrule(lr){3-4}\cmidrule(lr){5-6}\cmidrule(lr){7-8}
added to \bptt{} & seeds & \bptt{} & $K{=}1$ & repaired & broken & ER 800 & deg.\ 17 \\
\midrule
progressive loss, $\alpha{=}0.5$ & 5/23 & 1 & 0.062 & 2 & 1 & $+$0.034 (14+/5$-$) & $+$0.055 (13+/6$-$) \\
random initial state & 1/23 & 0.062 & 1 & 5 & 0 & $-$0.043 (5+/15$-$) & $-$0.067 (5+/15$-$) \\
\bottomrule
\end{tabular}
\end{table*}

\paragraph{The literature's remedies.} The progressive loss of \citet{bansal2022overthinking} adds, with weight $\alpha{=}0.5$, a rollout whose first $n_0$ applications run without gradient and next $k_0$ carry it, $n_0$ and $k_0$ drawn at random. Random initial states \citep{anil2022path} start half of each batch from Gaussian noise. They repair five seeds, break none and lower node accuracy on the band by 0.04 to 0.07 (15 of 23 seeds, $p=0.041$; Table~\ref{tab:remedies}).

\FloatBarrier
\section{Truncation}
\label{app:truncation}
\begin{table*}[!htbp]
\caption{Top: the truncation axis on the ER cells, seeds 0 to 7 of Table~\ref{tab:headtohead}; $K{=}1$ wins every paired comparison in 8/8 seeds, and $K{=}8$ gives the same results as \bptt{} in every seed. Bottom: the learning-rate control, seeds 0 to 5; ``wins'' counts seeds in which $K{=}1$ beats $K{=}2$ at the same rate, and the last two columns give the ratio of mean pre-clip gradient norms and the largest percentage of steps on which clipping fired in any run.}
\label{tab:trunc}
\centering\footnotesize\setlength{\tabcolsep}{4pt}
\begin{tabular}{lrrrrrrr}
\toprule
& \multicolumn{5}{c}{node accuracy, mean of 8 seeds} & \multicolumn{2}{c}{$K=1$ wins} \\
\cmidrule(lr){2-6}\cmidrule(lr){7-8}
test set & $K=1$ & $K=2$ & $K=4$ & $K=8$ & BPTT & vs $K=2$ & vs BPTT \\
\midrule
ER 800 & 0.854 & 0.652 & 0.662 & 0.688 & 0.688 & 8/8 & 8/8 \\
ER 800, deg.\ 17 & 0.837 & 0.537 & 0.555 & 0.604 & 0.604 & 8/8 & 8/8 \\
\bottomrule
\end{tabular}\\[5pt]
\begin{tabular}{lrrrrrr}
\toprule
& \multicolumn{2}{c}{ER 800 node acc.} & \multicolumn{2}{c}{$K=1$ wins (of 6)} & grad.\ norm & steps \\
\cmidrule(lr){2-3}\cmidrule(lr){4-5}
learning rate & $K=1$ & $K=2$ & ER 800 & deg.\ 17 & ratio, $K{=}1/K{=}2$ & clipped (\%) \\
\midrule
$1\cdot 10^{-3}$ & 0.816 & 0.696 & 6/6 & 6/6 & 0.83 & 0.2 \\
$2\cdot 10^{-3}$ & 0.844 & 0.628 & 6/6 & 6/6 & 0.93 & 0.2 \\
$4\cdot 10^{-3}$ & 0.725 & 0.552 & 5/6 & 5/6 & 1.07 & 0.1 \\
\bottomrule
\end{tabular}
\end{table*}
Does the break at $K{=}1$ depend on the window or on the learning rate? Detaching the state between applications is the $K{=}1$ end of truncated backpropagation. The progressive loss of \citet{bansal2022overthinking} backpropagates through a window of random length that starts at a random iteration, and \citet{geiping2025recurrentdepth} backpropagate through the last eight iterations. We swept $K\in\{2,4,8\}$ at eight seeds on the ER cells (Table~\ref{tab:trunc}); $K{=}1$ and full \bptt{} are the same seeds' stored runs. $K{=}1$ beats every other setting in 8 of 8 seeds at both cells. Among $K\ge 2$ and full \bptt{} the smallest pairwise sign-test $p$ is 0.375, so a residual slope of up to 0.07 node accuracy is not resolvable at eight seeds. $K{=}8$ gives the same results as \bptt{} in every seed; the two differ only on batches that hold a training graph more than eight hops deep. On the deep blob family, $K{=}2$ fails in 7/23 seeds against \bptt{}'s 6/23 ($p=1$, exact McNemar test on the same seeds) and \Kone{}'s 0/23 ($p=0.016$): one boundary crossing permits failure in this sample ($K{=}4$ and $8$ were not run there). The seven $K{=}2$ failures have the radius pattern of the \bptt{} floods: in four the radius grows no further after 16 cliques, and in three it grows to 32 cliques and falls short of the largest distance at 64. We did not read their per-node timing. Our windows start at fixed steps and every rollout starts from $h=0$ at step 0. The random start that \citet{bansal2022overthinking} credit with discouraging counting is tested with their progressive loss in \S\ref{sec:remedies}, where 5 of 23 seeds still flood.

Because $K$ changes how many \texttt{backward} calls accumulate into one optimizer step, the discontinuity could have been an effective-learning-rate artifact. A control specified in advance crossed $K\in\{1,2\}$ with three learning rates at six seeds (Table~\ref{tab:trunc}, bottom). $K{=}1$ wins at every rate, the best $K{=}2$ rate does not recover the gap, pre-clip gradient norms differ by 7\% at the operating rate with $K{=}1$'s the smaller, and clipping fires on about 0.2\% of steps or fewer. Read with \S\ref{sec:trigger}, the discontinuity says what a gradient across one application boundary can build on a tail input; it does not say which rule executes BFS better, since with the input fixed the two are identical everywhere.

\FloatBarrier
\begin{table*}[!htbp]
\caption{Sample standard deviations across the 23 paired head-to-head seeds. Means and paired sign counts are in Table~\ref{tab:headtohead}. Radius is in hops; accuracy is a fraction. The graph samples are fixed.}
\label{tab:dispersion}
\centering\footnotesize
\begin{tabular}{lrrrrrr}
\toprule
& \multicolumn{2}{c}{node accuracy} & \multicolumn{2}{c}{graph accuracy} & \multicolumn{2}{c}{radius} \\
\cmidrule(lr){2-3}\cmidrule(lr){4-5}\cmidrule(lr){6-7}
test set & BPTT & $K=1$ & BPTT & $K=1$ & BPTT & $K=1$ \\
\midrule
ER 16 & 0.000 & 0.000 & 0.000 & 0.000 & 0.000 & 0.000 \\
ER 80 & 0.003 & 0.003 & 0.018 & 0.018 & 0.000 & 0.000 \\
ER 160 & 0.038 & 0.042 & 0.120 & 0.147 & 0.864 & 1.033 \\
ER 800 & 0.107 & 0.192 & 0.101 & 0.261 & 0.507 & 2.130 \\
Delaunay 16 & 0.000 & 0.000 & 0.000 & 0.000 & 0.000 & 0.000 \\
Delaunay 80 & 0.000 & 0.000 & 0.000 & 0.000 & 0.000 & 0.000 \\
Delaunay 160 & 0.000 & 0.000 & 0.000 & 0.000 & 0.000 & 0.000 \\
Delaunay 800 & 0.000 & 0.000 & 0.000 & 0.000 & 0.000 & 0.000 \\
ER 800, target deg. 8 & 0.003 & 0.003 & 0.017 & 0.017 & 0.000 & 0.000 \\
ER 800, target deg. 12 & 0.160 & 0.171 & 0.449 & 0.435 & 0.834 & 0.846 \\
ER 800, target deg. 16 & 0.193 & 0.288 & 0.052 & 0.420 & 0.507 & 1.161 \\
ER 800, target deg. 24 & 0.011 & 0.216 & 0.000 & 0.175 & 0.000 & 0.647 \\
ER 800, target deg. 32 & 0.001 & 0.027 & 0.000 & 0.000 & 0.000 & 0.000 \\
\bottomrule
\end{tabular}
\end{table*}
\FloatBarrier
\section{Compute}
\label{app:compute}
The reported experiments contain \PTTrainingRuns{} completed GPU training records, including repeated seeds and reruns. These records are not independent models. Each uses 12,000 optimizer steps at batch 16 on a T4 GPU. The logs record about \PTTrainingHours{} GPU-hours of training in total, excluding evaluation and inference probes. These times are recorded to one decimal place in minutes. The benchmark's own models were attempted in 40 CPU training runs; 39 completed and one RecGNN run reached its time limit.

Every GPU training record contains all eight scheduled loss reports. From optimizer step 3,000 onward, each logged minibatch loss prints as 0.0000, rounded to four decimal places. These sampled losses support a good training fit. They do not check every training hint for every model. The separate full-pool check in Table~\ref{tab:density} verifies every hint on all 6,000 graphs for the six flooding GRU models that establish the paper's counterexample. The supplementary analysis recomputes the run counts, loss coverage and total logged training time from the per-run records.

\begin{table}[!htbp]
\caption{Recorded peak allocated GPU memory in a separate training-memory stress probe, batch 16 on a T4. $L$ counts the initial step; each sweep executes $L-1$ applications. BPTT's slope is between the first and last completed lengths; OOM is the first tested length that exhausted memory. $K{=}1$ has the listed peak at every tested power of two from $L=4$ through 8,192. These are controlled horizon sweeps, not peak memory measurements for every training run.}
\label{tab:memory}
\centering\footnotesize
\begin{tabular}{lrrr}
\toprule
& \multicolumn{2}{c}{BPTT} & $K{=}1$ \\
\cmidrule(lr){2-3}\cmidrule(lr){4-4}
family & MB/appl. & OOM at $L$ & GB, all $L$ \\
\midrule
Delaunay $n=800$ & 284 & 64 & 0.39 \\
grid $4\times 256$ & 273 & 64 & 0.36 \\
blob $16\times 64$ & 569 & 32 & 0.91 \\
\bottomrule
\end{tabular}
\end{table}
\FloatBarrier
\end{document}